\pdfoutput=1

\documentclass[11pt]{article}

\usepackage[preprint]{acl}

\usepackage{times}
\usepackage{latexsym}

\usepackage[T1]{fontenc}

\usepackage[utf8]{inputenc}

\usepackage{microtype}

\usepackage{inconsolata}

\usepackage{multicol}
\usepackage{booktabs}
\usepackage{makecell}
\usepackage{amsmath, amssymb}
\usepackage{multirow}
\usepackage{mathtools}
\usepackage{xcolor, colortbl}
\usepackage{enumitem}
\usepackage{float}
\usepackage{graphicx}
\usepackage{upgreek}
\usepackage{seqsplit}
\usepackage{color,soul}
\usepackage{arydshln}
\usepackage{placeins}
\usepackage{pifont}
\usepackage{verbatim}
\usepackage{bbm}
\usepackage{array}
\usepackage[table]{xcolor}
\usepackage[french,bordercolor=none,backgroundcolor=pink,linecolor=none,textsize=footnotesize,textwidth=0.75in,shadow]{todonotes}

\usepackage{amsmath,amsfonts,bm}

\def\eqref#1{equation~\ref{#1}}

\def\1{\bm{1}}

\def\vq{{\bm{q}}}
\def\vr{{\bm{r}}}
\def\vs{{\bm{s}}}

\def\vx{{\bm{x}}}
\def\vy{{\bm{y}}}

\DeclareMathAlphabet{\mathsfit}{\encodingdefault}{\sfdefault}{m}{sl}
\SetMathAlphabet{\mathsfit}{bold}{\encodingdefault}{\sfdefault}{bx}{n}

\usepackage{caption}
\usepackage{subcaption}
\definecolor{myred}{RGB}{205,33,42}
\definecolor{darkorchid}{rgb}{0.6, 0.2, 0.8}
\definecolor{darkmidnightblue}{rgb}{0.0, 0.2, 0.4}
\definecolor{darkspringgreen}{rgb}{0.09, 0.45, 0.27}
\definecolor{myblue}{rgb}{0.0, 0.5, 1.0}

\title{Adaptive Multi-Agent Response Refinement in Conversational Systems}

\author{
    Soyeong Jeong$^1$\thanks{\hspace{0.01cm} Work done during internship at Amazon.}
    \; Aparna Elangovan$^3$\thanks{\hspace{0.01cm} Work done while at Amazon.}
    \; Emine Yilmaz$^{2, 4}$
    \; Oleg Rokhlenko$^2$
    \\
    KAIST$^{1}$ \;\; Amazon$^{2}$ \;\; Collate$^{3}$ \;\; University College London$^{4}$\\
   \texttt{starsuzi@kaist.ac.kr}\\
}

\begin{document}
\maketitle
    
\begin{abstract}
Large Language Models (LLMs) have demonstrated remarkable success in conversational systems by generating human-like responses. However, they can fall short, especially when required to account for personalization or specific knowledge. In real-life settings, it is impractical to rely on users to detect these errors and request a new response. One way to address this problem is to refine the response before returning it to the user. While existing approaches focus on refining responses within a single LLM, this method struggles to consider diverse aspects needed for effective conversations. In this work, we propose refining responses through a multi-agent framework, where each agent is assigned a specific role for each aspect. We focus on three key aspects crucial to conversational quality: factuality, personalization, and coherence. Each agent is responsible for reviewing and refining one of these aspects, and their feedback is then merged to improve the overall response. To enhance collaboration among them, we introduce a dynamic communication strategy. Instead of following a fixed sequence of agents, our approach adaptively selects and coordinates the most relevant agents based on the specific requirements of each query. We validate our framework on challenging conversational datasets, demonstrating that ours significantly outperforms relevant baselines, particularly in tasks involving knowledge or user's persona, or both.
\end{abstract}

\section{Introduction}
In recent years, Large Language Models (LLMs) have demonstrated remarkable performance across a broad spectrum of NLP tasks, primarily due to their ability to generate coherent and contextually relevant responses, powered by extensive training on diverse data~\citep{Liu2024FromLT, llm_conv_survey1}. However, LLMs do not always produce satisfactory responses on the first attempt~\citep{DBLP:conf/acl/ChiesurinDCEPRK23, platypus2023}, and this issue becomes more evident in multi-turn conversational settings, where models must not only interpret the user's current query but also consider the entire conversational history, including dependencies, ambiguities, and co-references~\citep{excord, DBLP:conf/emnlp/JangL23a, DBLP:conf/coling/GanPY24}. These difficulties are particularly pronounced in complex, personalized conversations that demand both user's persona alignment and factual accuracy~\citep{wikichat, synthetic_personachat}. In such situations, users frequently have to prompt the LLM to correct its responses, which can interrupt the conversation flow and diminish the overall user experience~\citep{self-refine, DBLP:conf/emnlp/DengLC0LC23,kim-etal-2024-commonsense}. Thus, LLMs should proactively refine their inaccurate responses, even without explicit user requests for correction.

Several approaches have been explored for refining responses using a single agent, such as Self-Refine~\citep{self-refine}, where a single agent handles an entire refinement process, generating feedback across multiple aspects and iteratively improving its outputs based on it. However, relying solely on a single agent can be suboptimal, as the agent may become overly confident in the initial output, leading to bias~\citep{mad}, which can be amplified through repeated iterations~\citep{DBLP:conf/acl/XuZZP0024, DBLP:conf/iclr/0009CMZYSZ24}. This issue could be particularly critical for multi-turn conversational tasks, where errors in earlier turns can propagate to subsequent turns~\citep{DBLP:conf/eacl/JeongBHP23}.

\begin{figure*}
    \centering
    \includegraphics[width=0.975\linewidth]{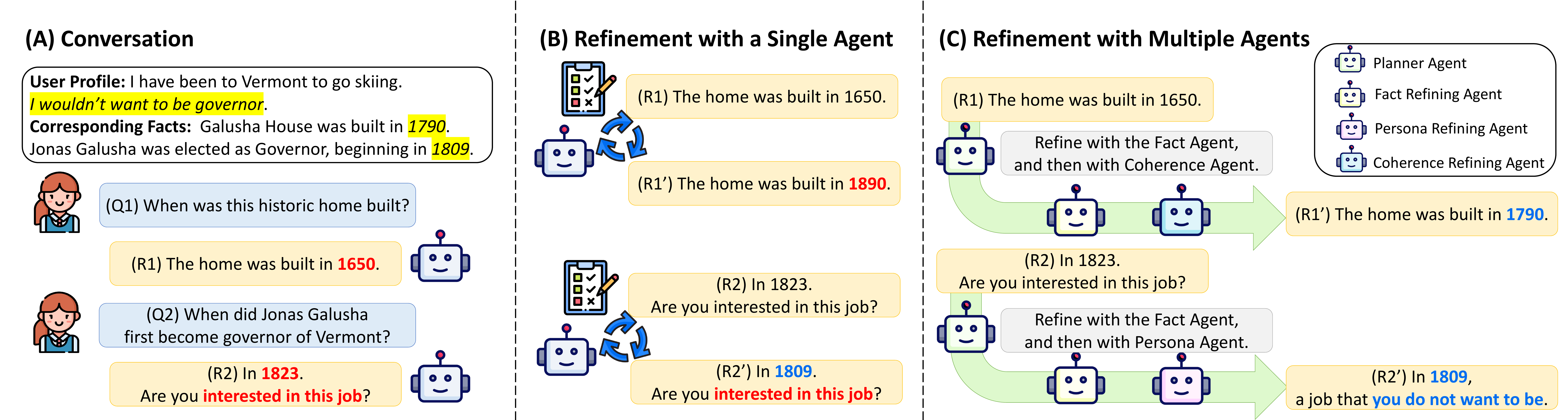}
    \vspace{-0.05in}
    \caption{(A) A customized conversation which requires alignment with both the user profile and specific fact. The responding agent fails to provide accurate information in R1 and also fails to align with the user's profile in R2, as indicated in \textcolor{red}{red}. (B) In a single-agent-based refinement, one agent manages all the refinement procedures. While some misalignments can be addressed, as indicated in \textcolor{blue}{blue}, the result may still be suboptimal. (C) Our multi-agent-based refinement, where multiple agents, each focusing on a different aspect, collaboratively refine responses based on the set and sequence determined by a planner agent.}
    \label{fig:concept}
    \vspace{-0.1in}
\end{figure*}

A more effective approach is to incorporate multiple agents, each with a specific perspective. This method takes advantage of LLMs' ability to tackle complex tasks by decomposing them into simpler subtasks and enabling collaborative problem-solving~\citep{autogen, DBLP:journals/corr/abs-2402-01680, multi_agent_debate}. Additionally, the multi-agent framework enables for integration of a broader range of tools and knowledge sources~\citep{DBLP:journals/corr/abs-2401-07324, DBLP:journals/corr/abs-2402-14034, multiagent_survey1}, ensuring that each agent can be specifically tailored to meet the diverse needs of each task. While multi-agent frameworks have shown great promise in fields such as human behavior simulation, economic theories, and more, their potential for response refinement in conversational systems remains largely underexplored. Figure~\ref{fig:concept} (A) exemplifies a conversation where a response should accurately reflect the user profile and specific knowledge. In such setting, employing multiple refining agents with specialized perspectives can enhance conversation quality. Therefore, in this work, we introduce multiple refining agents that refine responses when the initial outputs are incorrect, each focused on a distinct perspective: user persona alignment (persona-refining agent), factual grounding to mitigate hallucinations (fact-refining agent), and logical coherence with previous conversational turns (coherence-refining agent).

The remaining challenge, then, is how to enhance the overall quality of the conversation to deliver the most satisfactory refined response to the user, utilizing our three refining agents. To achieve this, the agents should collaborate differently for each query, as each query, even within the same conversation, may require a different focus. For example, as shown in Figure~\ref{fig:concept} (A), some queries require factual knowledge, while some queries also demand an understanding of the user's profile. Therefore, the set of agents deployed should vary depending on the query's specific needs. Additionally, the optimal sequence in which the agents are engaged may vary, as the focus and priorities of each query can differ. For example, a query requiring immediate factual verification might first engage the fact-refining agent to correct inaccuracies, followed by other agents as necessary. Therefore, we further propose a dynamic refinement process that adjusts both the set and the sequence of refining agents for each query, by introducing a novel planner agent that generates a sequence of required refining agents tailored to the query, along with justifications for each decision. Each refining agent then refines the response sequentially, following the sequence and referencing the justifications provided by the planner agent. Note that all agents are based on unsupervised LLMs, each instantiated with a prompt specifying a specific role. We refer to this framework as \textbf{M}ulti-\textbf{A}gent \textbf{R}efinement with \textbf{A}daptive agent selection (MARA), which is illustrated in Figure~\ref{fig:concept} (C).

We validate the efficacy of our framework on challenging conversational datasets that involve user persona integration, require specific factual knowledge, or both. The results show that MARA significantly outperforms relevant baselines, indicating that refining along the three aspects, persona, factuality, and coherence, is beneficial. Moreover, our analyses highlight the importance of dynamically selecting appropriate refinement strategies tailored to the specific needs of each query in diverse conversational contexts.

\section{Related Work}
\vspace{-0.05in}

\noindent \textbf{LLMs in Conversation Systems.}
Recent Large Language Models (LLMs) have demonstrated remarkable performance across a range of NLP tasks~\citep{gpt4, claude, llama}, even without additional training, largely due to their massive pretraining on diverse datasets~\citep{DBLP:conf/emnlp/MinLHALHZ22, cot}. Their success is also remarkable in conversational tasks~\citep{DBLP:conf/emnlp/ChaeSOKKYLKY23, Liu2024FromLT, llm_conv_survey1}, as LLMs are able to generate human-like responses~\citep{DBLP:conf/acl/LeeHPP023, DBLP:journals/corr/abs-2407-02397, DBLP:conf/emnlp/DengLC0LC23}. 
However, LLMs occasionally produce unsatisfactory responses, particularly when they fail to consider a user's profile~\citep{synthetic_personachat}, specific knowledge~\citep{wikichat}, or are not coherent with the previous turns~\citep{DBLP:journals/tois/HuangZG20, DBLP:conf/naacl/OuLLTZZG24}.

\vspace{0.075in}
\noindent \textbf{Verification and Refinement with LLMs.}
To address the issue of unsatisfactory responses, several studies have explored verification and refinement approaches. While some research involves the use of additional refinement models,~\citep{DBLP:conf/emnlp/BaekJKPH23, DBLP:conf/naacl/XuDFJZLWLF24}, more recent studies emphasize the capability of LLMs to verify and refine their own responses without the need for further training~\citep{DBLP:conf/nips/ShinnCGNY23, react, DBLP:journals/corr/abs-2406-15673, DBLP:conf/iclr/GouSGSYDC24,chen-etal-2024-reconcile}. Specifically,~\citet{self-refine} proposed a method where a single agent LLM generates feedback on multiple aspects of its own response. 
However, single-agent-based refinements can be less effective, as the overall performance is constrained by the capacity of a single agent, and once an agent becomes confident in its outputs, it may struggle to further generate novel thoughts~\citep{mad,DBLP:conf/acl/XuZZP0024, DBLP:conf/iclr/0009CMZYSZ24}.

\vspace{0.075in}
\noindent \textbf{Multi-Agent LLMs.}
Along with the powerful capabilities of an LLM, its performance can be further enhanced when multiple LLMs are involved by collaborating or debating each other, particularly on complex problems that require diverse perspectives from specialized LLM agents~\citep{DBLP:conf/nips/LiHIKG23, autogen, DBLP:journals/corr/abs-2402-01680, camel}. Multi-agent LLM framework has recently been widely applied to various tasks, including software development~\citep{DBLP:conf/iclr/HongZCZCWZWYLZR24,DBLP:conf/acl/QianLLCDL0CSCXL24}, model evaluation~\citep{chateval}, research topic generation~\citep{ResearchAgent}, diagnostic consultations~\citep{diaggpt,mdagent}, recommendation~\citep{DBLP:journals/corr/abs-2402-01135,Spurlock2024ChatGPTFC}, and reasoning improvement~\citep{Chen2024MAgICoReMI}.
However, the use of multiple agents for refining conversational turns remains underexplored.

\section{Method}
We introduce MARA, a multi-agent framework that dynamically refines conversational responses.

\subsection{Preliminaries}
We begin by formally defining an LLM, specifically in the context of a multi-turn conversational setting.

\vspace{0.075in}
\noindent \textbf{Large Language Models.}
Let us first define LLM as \texttt{LLM}, which takes an input sequence of tokens $\vx$ and generates an output sequence of tokens $\vy$. This process can be represented as $\vy = \texttt{LLM}(\mathcal{P}(\vx))$, where the prompt template $\mathcal{P}$ incorporates additional context or instructions that guide the LLM's behavior. Specifically, $\mathcal{P}$ can be used for role assignment, ensuring that the LLM adopts a particular role or performs a specific task.

\vspace{0.075in}
\noindent \textbf{Conversation with an LLMs.}
In a multi-turn conversational setting, the LLM, acting as a responding agent, generates a response $\vr^i$ for the $i$-th turn based on the current query $\vq^i$ and the preceding conversational context, which can be represented as: $\vr^i = \texttt{LLM}(\mathcal{P}_{\texttt{respond}}(\vq^i, \vq^{i-1}, \vr^{i-1}, ..., \vq^1, \vr^1))$, where $\vq^i$ is the user's query at the $i$-th turn, and $\vr^i$ is the response generated by the \texttt{LLM} which operates under the prompt template $\mathcal{P}_{\texttt{respond}}$. However, the quality of initial responses from the LLM may be suboptimal in customized, realistic conversations, potentially overlooking the user's profile, missing specific knowledge, or failing to maintain coherence, and such errors in earlier conversational turns can cumulatively impact subsequent turns.

\subsection{Multi-Agent Response Refinement}
We now turn to our primary focus of further refining the initial response. To achieve this, we define three specialized refining agents, each responsible for enhancing different aspects of the response.

\vspace{0.075in}
\noindent \textbf{Single-Agent Response Refinement.}
As an initial response $\vr$ may be inaccurate or unsatisfactory, the goal of response refinement is to further improve its quality. This can be achieved by using an LLM operating as a refining agent, represented as: $\vr_{\texttt{refine}} = \texttt{LLM}(\mathcal{P}_{\texttt{refine}}(\vr))$, where an LLM instantiated with $\mathcal{P}_{\texttt{refine}}$ refines an initial response $\vr$ to produce the improved response $\vr_{\texttt{refine}}$. However, relying on a single agent to assess and refine multiple aspects can be ineffective, as it may struggle to holistically address the diverse factors essential for a high-quality, customized conversation. To overcome these limitations, we propose a multi-agent framework in which specialized refining agents collaborate to enhance response quality.

\vspace{0.075in}
\noindent \textbf{Multiple Refining Agents for Conversation.}
In a realistic yet challenging conversational scenario, responses must be contextually relevant, aligned with user preferences, and factually reliable. To meet these requirements, responses must be refined across multiple dimensions, ensuring factual correctness, alignment with the user’s persona, and coherence across multiple conversational turns. Here, instead of addressing all these aspects with a single agent, we introduce three specialized refining agents that extend the general refining agent, $\texttt{LLM}(\mathcal{P}_{\texttt{refine}}(\vr))$, each addressing a distinct aspect of response quality.
Specifically, the fact-refining agent generates the refined response, $\vr_{\texttt{fact-refine}}$, that ensures factual accuracy, $\vr_{\texttt{persona-refine}}$ that aligns responses with the user's profile, and $\vr_{\texttt{coherence-refine}}$ that maintains coherence throughout conversation, where each $\texttt{LLM}$ is instantiated with its respective refining role template: $\mathcal{P}_{\texttt{fact-refine}}$, $\mathcal{P}_{\texttt{persona-refine}}$, and $\mathcal{P}_{\texttt{coherence-refine}}$. Having defined the role of each refining agent, the next key challenge lies in how these agents should collaborate to collectively refine the response.

\subsection{Communication Strategy among Agents}
To enable effective collaboration among multiple agents, we explore various communication strategies and propose a dynamic approach to optimize the refinement process.

\vspace{0.075in}
\noindent \textbf{Simultaneous Communication.}
As an initial approach to agent collaboration, we introduce a simultaneous communication strategy, where all refining agents independently refine the initial response, and their refined outputs are then passed to a finalizer agent instantiated with $\mathcal{P}_\texttt{finalizer}$, which aggregates the individual refinements into a single unified response. Formally, this process is denoted as: $\texttt{LLM}(\mathcal{P_{\texttt{finalizer}}}(\vr_{\texttt{fact-refine}},$ $\vr_{\texttt{persona-refine}},$ $\vr_{\texttt{coherence-refine}}))$. However, this simultaneous approach always necessitates an additional finalizer agent, and the overall quality of the final output may heavily depend on the capabilities of this finalizer agent.

\vspace{0.075in}
\noindent \textbf{Sequential Communication.}
To address potential limitations of the simultaneous approach, we introduce a sequential refinement process where each agent builds upon the response refined by the previous agent. 
Specifically, given an initial response $\vr$, it is sequentially refined by a series of agents denoted as $\vs = [{\texttt{LLM}(\mathcal{P}_{\texttt{refine}_k}}(\vr))]_{k=1}^{n}$, where each refining agent is instantiated with a role-specific prompt template $\mathcal{P}_{\texttt{refine}_k}$, selected from the set of refinement templates: $\{\mathcal{P}_{\texttt{fact-refine}},$ $\mathcal{P}_{\texttt{persona-refine}},$ $\mathcal{P}_{\texttt{coherence-refine}}\}$.
Here, each refining agent receives the response refined by the preceding agent, $\vr_{\texttt{refine}_{k-1}}$, as input, and applies its specific prompt template, $\mathcal{P}_{\texttt{refine}_k}$, to generate the next refined response, $\vr_{\texttt{refine}_{k}}$. The sequential process continues through the sequence of agents until obtaining the final refined response, $\vr_{\texttt{refine}_n}$. 
Yet, it may require a different combination and ordering of refining agents based on the specific context, but the optimal sequence of refining agents may vary depending on the conversational context.

\vspace{0.075in}
\noindent \textbf{Dynamic Sequential Communication.}
Therefore, to further adaptively operate the response refinement process, we introduce a dynamic strategy, where a planner agent selects the most suitable sequence of refining agents along with justifications for its decisions, adapting to the needs of each query. Specifically, the planner agent determines the sequence of refining agents required for each query, instantiated by a template $\mathcal{P}_{\texttt{planner}}$.
Formally, given a query $\vq$ and an initial response $\vr$, the planner agent outputs the sequence of refining agents as follows: $\vs_{\texttt{planner}} = \texttt{LLM}(\mathcal{P}_{\texttt{planner}}(\vq, \vr))$. Once $\vs_{\texttt{planner}}$ is determined, the refining agents sequentially refine the response, with each agent taking as input the refined response from the preceding agent in the sequence. Additionally, each agent also receives the planner’s justifications, allowing each agent to understand its role in the sequence and collaborate effectively.

\section{Experimental Setups}
In this section, we describe the experimental setup, leaving further details in Appendix~\ref{appendix:experimental_setup}.

\begin{table*}[t!]
\caption{\small Results on three datasets, using Claude as the base LLM, with statistically significant best results highlighted in \textbf{bold}. Additionally, MARA$^{\ast}$ is a variant of MARA where the fact-refining agent uses the same LLM as the responding agent.}
\vspace{-0.1in}
\label{tab:main}
\small
\centering
\resizebox{\textwidth}{!}{
\renewcommand{\arraystretch}{1.25}
\setlength{\tabcolsep}{1pt}
\begin{tabular}{l ccccc ccccc ccccc}
\toprule

& \multicolumn{5}{c}{\bf PersonaChat} & \multicolumn{5}{c}{\bf INSCIT} & \multicolumn{5}{c}{\bf FoCus} \\
\cmidrule(l{2pt}r{2pt}){2-6} \cmidrule(l{2pt}r{2pt}){7-11} \cmidrule(l{2pt}r{2pt}){12-16}

\textbf{Methods} & \textbf{Coh. (1-3)} & \textbf{Grd. (0-1)} & \textbf{Nat. (1-3)} & \textbf{Eng. (1-3)} & \textbf{Overall} & \textbf{Coh. (1-3)} & \textbf{Grd. (0-1)} & \textbf{Nat. (1-3)} & \textbf{Eng. (1-3)} & \textbf{Overall} & \textbf{Coh. (1-3)} & \textbf{Grd. (0-1)} & \textbf{Nat. (1-3)} & \textbf{Eng. (1-3)} & \textbf{Overall} \\

\midrule
\midrule

\textbf{No Refine} & 1.87 \scriptsize{± 0.01} & 0.42 \scriptsize{± 0.00} & 1.55 \scriptsize{± 0.01} & 1.55 \scriptsize{± 0.00} & \cellcolor{gray!20} 34.98  & 2.53 \scriptsize{± 0.02} & 0.52 \scriptsize{± 0.02} & \textbf{2.36} \scriptsize{± 0.01} & 2.36 \scriptsize{± 0.01} & \cellcolor{gray!20} 65.07 & 2.39 \scriptsize{± 0.02} & 0.49 \scriptsize{± 0.02} & 2.02 \scriptsize{± 0.01} & 2.15 \scriptsize{± 0.02} & \cellcolor{gray!20} 56.71 \\

\noalign{\vskip 0.25ex}\cdashline{1-16}\noalign{\vskip 0.75ex}
			
\textbf{Self-Refine}  & 2.26 \scriptsize{± 0.01} & 0.52 \scriptsize{± 0.01} & 2.07 \scriptsize{± 0.01} & 2.30 \scriptsize{± 0.03} & \cellcolor{gray!20} 58.41 & 2.31 \scriptsize{± 0.01} & 0.45 \scriptsize{± 0.00} & 2.19 \scriptsize{± 0.01} & 2.24 \scriptsize{± 0.01} & \cellcolor{gray!20} 57.83 & 2.10 \scriptsize{± 0.01} & 0.37 \scriptsize{± 0.01} & 1.87 \scriptsize{± 0.00} & 2.05 \scriptsize{± 0.02} & \cellcolor{gray!20} 47.07 \\

\textbf{SPP}  & 1.98 \scriptsize{± 0.02} & 0.44 \scriptsize{± 0.01} & 1.72 \scriptsize{± 0.02} & 2.07 \scriptsize{± 0.01} &  \cellcolor{gray!20} 45.55 & 2.52 \scriptsize{± 0.01} & 0.49 \scriptsize{± 0.01} & 2.29 \scriptsize{± 0.01} & 2.47 \scriptsize{± 0.02} & \cellcolor{gray!20} 65.74 & 2.44 \scriptsize{± 0.00} & 0.47 \scriptsize{± 0.01} & 1.99 \scriptsize{± 0.02} & 2.46 \scriptsize{± 0.02} & \cellcolor{gray!20} 60.47 \\

\noalign{\vskip 0.25ex}\cdashline{1-16}\noalign{\vskip 0.75ex}

\textbf{LLMvLLM} & 1.24 \scriptsize{± 0.01} & 0.26 \scriptsize{± 0.00} & 1.07 \scriptsize{± 0.00} &  1.17 \scriptsize{± 0.01} &  \cellcolor{gray!20} 12.57 & 2.24  \scriptsize{± 0.00} &  0.47 \scriptsize{± 0.00} &  1.79 \scriptsize{± 0.01} &  2.15 \scriptsize{± 0.00} & \cellcolor{gray!20} 51.59 & 1.77 \scriptsize{± 0.01} &  0.32 \scriptsize{± 0.00} &  1.34 \scriptsize{± 0.00} & 1.71 \scriptsize{± 0.01} &  \cellcolor{gray!20} 30.92 \\

\textbf{MADR} & 1.59 \scriptsize{± 0.03} & 0.28 \scriptsize{± 0.01} & 1.38 \scriptsize{± 0.03} & 1.32 \scriptsize{± 0.01} &  \cellcolor{gray!20} 23.21 & 2.29 \scriptsize{± 0.03} & 0.46 \scriptsize{± 0.01} & 2.12 \scriptsize{± 0.04} & 1.89 \scriptsize{± 0.04} & \cellcolor{gray!20} 52.89 & 1.93 \scriptsize{± 0.03} & 0.32 \scriptsize{± 0.01} & 1.68 \scriptsize{± 0.04} & 1.54 \scriptsize{± 0.03} & \cellcolor{gray!20} 34.71 \\

\textbf{MultiDebate}  & 1.78 \scriptsize{± 0.01} &  0.36 \scriptsize{± 0.01} & 1.58 \scriptsize{± 0.00} & 1.68 \scriptsize{± 0.00} & \cellcolor{gray!20} 34.70 &  2.44 \scriptsize{± 0.00} &  0.53 \scriptsize{± 0.01} & 2.27  \scriptsize{± 0.01} &  2.18 \scriptsize{± 0.01} & \cellcolor{gray!20} 61.70 & 2.31 \scriptsize{± 0.00} & 0.51  \scriptsize{± 0.00} & 1.93 \scriptsize{± 0.00} & 2.12 \scriptsize{± 0.00} & \cellcolor{gray!20} 54.81 \\

\noalign{\vskip 0.25ex}\cdashline{1-16}\noalign{\vskip 0.75ex}

\textbf{MARA$^{\ast}$} &  2.28  \scriptsize{± 0.01} &   0.56 \scriptsize{± 0.00} & 1.91 \scriptsize{± 0.01} & 2.52 \scriptsize{± 0.00} & \cellcolor{gray!20} 60.24 & 2.54  \scriptsize{± 0.01} & 0.55 \scriptsize{± 0.01} & 2.10 \scriptsize{± 0.01} & 2.70 \scriptsize{± 0.00} & \cellcolor{gray!20} 67.79 & 2.56 \scriptsize{± 0.01} & 0.58 \scriptsize{± 0.01} & 1.98 \scriptsize{± 0.01} & 2.81 \scriptsize{± 0.02} & \cellcolor{gray!20} 68.77 \\

\rowcolor{blue!5} 
\textbf{MARA (Ours)} & \textbf{2.32} \scriptsize{± 0.01} & \textbf{0.56} \scriptsize{± 0.01} & \textbf{1.98} \scriptsize{± 0.01} & \textbf{2.54} \scriptsize{± 0.06} & \cellcolor{gray!20} \textbf{62.00} & \textbf{2.62} \scriptsize{± 0.02} & \textbf{0.59} \scriptsize{± 0.01} & 2.26 \scriptsize{± 0.03} & \textbf{2.74} \scriptsize{± 0.07} & \cellcolor{gray!20} \textbf{72.61} & \textbf{2.67} \scriptsize{± 0.00} & \textbf{0.65} \scriptsize{± 0.01} & \textbf{2.15} \scriptsize{± 0.01} & \textbf{2.83} \scriptsize{± 0.00} &  \cellcolor{gray!20} \textbf{74.51} \\

\bottomrule

\end{tabular}
}
\vspace{-0.1in}
\end{table*}

\subsection{Datasets}
We validate MARA in realistic conversational settings, including queries requiring personalization and factual information.
\textbf{PersonaChat}~\citep{personachat} is designed to generate responses aligned with a user’s persona.
\textbf{INSCIT}~\citep{inscit} is an information seeking dataset where knowledge is grounded in Wikipedia.
\textbf{FoCus}~\citep{focus} requires both user-aligned and knowledge-grounded responses. Furthermore, we conduct evaluations on two other datasets: the role-playing dataset, \textbf{PRODIGy}~\citep{PRODIGy}, and the domain-specific, \textbf{Ubuntu Dialogue Corpus}~\citep{ubuntu}.

\subsection{Baselines and Our Model}
We compare MARA against six baseline models, starting with a baseline without refinement, two single-agent refinement approaches, and three multi-agent refinement approaches. 
\textbf{1) No Refine} generates response without any further refinement. 
\textbf{2) Self-Refine}~\citep{self-refine} is a single agent refinement baseline, which makes an LLM to generate feedback on 10 aspects of its own response and iteratively refines with the feedback. 
\textbf{3) SPP}~\citep{spp} is another single-agent refinement baseline, where a single LLM generates multiple self-constructed roles that collaborate within a single prompt. 
\textbf{4) LLMvLLM}~\citep{llmvllm} is a multi-agent refinement baseline where two agents engage in cross-examination to detect factual errors in generated response.
\textbf{5) MADR}~\citep{madr} is another multi-agent refinement baseline where two agents debate based on the predefined error types to generate fact checking explanations. 
\textbf{6) MultiDebate}~\citep{multidebate} is a multi-agent refinement baseline where multiple language model agents iteratively engage in a structured debate to improve factual accuracy and reasoning.
\textbf{7) MARA} is our proposed system, where the agents refine the response in the specified order, adaptively generated by a planner agent.

\subsection{Evaluation Metrics}
We evaluate the models using G-Eval~\citep{geval} to assess the quality of the refined responses. Following the setup in \citet{geval} for a conversational setting, we assess each refined response using four metrics: \textbf{1) Coherence (Coh.)} measures whether the conversation response logically follows the preceding context with a scale of 1 to 3, which is highly related to the effectiveness of the coherence-refining agent. 
\textbf{2) Groundedness (Grd.)} evaluates whether the response accurately incorporates the provided fact, with a scale of 0 to 1, which corresponds to the effectiveness of the fact-refining agent. 
\textbf{3) Naturalness (Nat.)} evaluates whether the response is natural with a scale of 1 to 3. 
\textbf{4) Engagingness (Eng.)} measures whether the response is engaging with a scale of 1 to 3, which mainly reflects the effectiveness of the persona-refining agent.
Furthermore, to provide a comprehensive assessment, we report the \textbf{5) Overall} score, which represents a scale-normalized average of the four evaluation metrics.

\subsection{Implementation Details}
We mainly use the Claude Sonnet 3 model~\citep{claude} as the base LLM for both our framework and the baselines. Furthermore, since a multi-agent framework offers the flexibility to assign different LLMs to each agent based on their role, for the fact-refining agent, we use another Claude Sonnet 3.5 model, as it offers improved capabilities in factual accuracy, which are critical for this particular role. Additionally, we report the performance of MARA$^{\ast}$, a variant of MARA in which the fact-refining agent uses Sonnet 3 instead, ensuring that all agents use the same model.
In order to assess the robustness of our proposed framework across diverse LLMs, we further use GPT-4o-mini~\citep{gpt4}, LLaMA 3.1 8B, and LLaMA 3.1 70B~\citep{llama}.
For each dataset, we sample 100 conversations, resulting in a total of 673 queries for the PersonaChat dataset, 506 queries for the INSCIT dataset, and 563 queries for the FoCus dataset. For the G-Eval metric, we use GPT-4o mini model~\citep{gpt4}, with the normalization steps.
We include the prompts used for MARA and G-Eval in Appendix~\ref{appendix:experimental_setup}.

\section{Experimental Results and Analyses}
We present the experimental results and analyses.

\subsection{Main Results}
Here, we present the overall results across various challenging yet realistic conversational scenarios.

\vspace{0.075in}
\noindent \textbf{Overall Results.}
We report the overall experimental results in Table~\ref{tab:main} with three different runs. As shown in the table, MARA consistently outperforms other models across diverse metrics, and the gaps between MARA and the other models are significantly substantial. 

To be more specific, we explore diverse conversational scenarios, particularly challenging ones requiring alignments with the user's persona (PersonaChat), factual grounding (INSCIT), and a combination of both (FoCus). In PersonaChat, which emphasizes responses that align with the user's profile, single-agent refinement methods such as Self-Refine and SPP outperform the baseline without refinement. This improvement is likely due to their focus on enhancing fluency and coherence, which aligns well with tasks requiring the incorporation of user interests and preferences across conversations, rather than managing specific factual content. 
However, in datasets that demand specific factual knowledge, such as INSCIT and FoCus, single-agent refinement becomes less effective and even results in performance degradation, particularly in groundedness scores. Notably, SPP outperforms Self-Refine, suggesting that generating multiple roles within a single agent can be beneficial for information-intensive queries.
Nevertheless, compared to the significantly improved performance of our MARA framework, generating multiple perspectives through distinct agents appears to be more effective.

\begin{figure}
    \centering
    \includegraphics[width=0.9\linewidth]{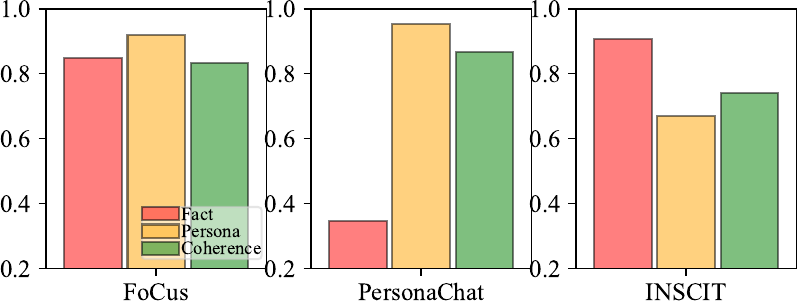}
    \vspace{-0.1in}
    \caption{Distribution of refining agents for three datasets.}
    \label{fig:agent_distribution_bar}
    \vspace{-0.18in}
\end{figure}

However, adopting a multi-agent framework does not necessarily guarantee improved performance, especially in challenging customized conversational tasks that require user persona understanding and factual knowledge. In fact, the significant performance gap between MARA and other multi-agent baselines highlights the crucial role of agent design in enhancing the effectiveness of multi-agent frameworks. In other words, compared to other multi-agent frameworks, which either focus only on predefined factual error types (MADR) or lack explicitly defined agent roles for targeted refinement (LLMvLLM and MultiDebate), these design choices may be less effective in challenging conversational scenarios that require multiple aspects.
In contrast, MARA consistently improves performance across all three conversational scenarios, demonstrating the effectiveness of our three-agent design, specifically tailored for user profile alignment, factual accuracy, and coherence. 

\begin{table}[t!]
\caption{\small Analyses on a planner agent, comparing performance with randomly or ideally assigned refining agents.}
\label{tab:planner}
\vspace{-0.075in}
\small
\centering
\resizebox{0.475\textwidth}{!}{
\begin{tabular}{lcccccc}
\toprule
\textbf{Planner Design} & \textbf{Coh.} & \textbf{Grd.} & \textbf{Nat.} & \textbf{Eng.} & \textbf{\#Agent}\\

\midrule
\midrule

\multirowcell{1}[-0.0ex][l]{\textbf{No Refine}}
& 2.21 & 0.40 & 1.88 & 2.05 & 1.0 \\

\noalign{\vskip 0.25ex}\cdashline{1-6}\noalign{\vskip 0.75ex}

\multirowcell{1}[-0.0ex][l]{\textbf{MARA w/ Random}}
& 2.42 & 0.47 & 1.95 & 2.64 & 2.0 \\

\multirowcell{1}[-0.0ex][l]{\textbf{MARA}}
& 2.54 & 0.58 & 2.07 & 2.76 & 4.4 \\

\multirowcell{1}[-0.0ex][l]{\textbf{MARA w/ Ideal}}
& \textbf{2.88} & \textbf{0.80} & \textbf{2.58} & \textbf{2.93} & 3.0 \\

\bottomrule
\end{tabular}
}
\vspace{-0.05in}
\end{table}

\vspace{0.075in}
\noindent \textbf{Effectiveness of our Planner Agent.}
Notably, in realistic conversational scenarios, different queries (even within the same conversation) focus on different aspects, thus requiring distinct sets of refining agents. To address this, we propose a planner agent that dynamically assigns refining agents based on the specific requirements of each query. To evaluate its effectiveness, we first analyze the distribution of refining agents across three different conversational settings. 
As shown in Figure~\ref{fig:agent_distribution_bar}, the distribution varies across different conversational datasets, indicating that the planner agent effectively adapts to diverse query requirements. Specifically, the persona-refining agent is predominantly used in datasets where responses must incorporate user preferences, while the fact-refining agent is more frequently required in knowledge-focused datasets. For the dataset requiring both aspects, both agents are allocated in similar proportions. 

Next, we further evaluate the effectiveness of the current design of our planner agent by comparing it against a random setting. As shown in Table~\ref{tab:planner}, the performance of a randomly assigned sequence of refining agents is lower than that of our planner-driven approach, indicating that the proposed planner effectively determines an optimal refinement sequence. However, even with randomly assigned agents, performance remains higher than the baseline without refinement, demonstrating the effectiveness of incorporating our three refining agents.
Furthermore, we explore the full potential of our planner agent by investigating how much further it can improve performance. To do so, we evaluate an ideal planner that selects the best-performing sequence. We report its performance by brute-forcing all possible combinations of sequences using our three refining agents and selecting the optimal sequence for each query\footnote{We report results for 20 conversations in Table~\ref{tab:planner}, as full brute-force computation is largely expensive.}. As shown in Table~\ref{tab:planner}, this ideal planner achieves the highest performance while requiring fewer accesses to the refining agents. These results validate the effectiveness of our dynamic agent allocation strategy and indicate that further advancements in planner agents could yield even greater performance improvements.

\begin{table}[t!]
\caption{\small Results on the role-play dataset (PRODIGy), using GPT as the base LLM. Best results highlighted in \textbf{bold}.}
\vspace{-0.1in}
\label{tab:roleplay}
\small
\centering
\resizebox{0.475\textwidth}{!}{
\renewcommand{\arraystretch}{1.0}
\begin{tabular}{l ccccc}
\toprule
\textbf{Methods} & \textbf{Coh.} & \textbf{Grd.} & \textbf{Nat.} & \textbf{Eng.} & \textbf{Overall}\\

\midrule
\midrule

{\textbf{No Refine}}
& 2.15  & 0.41  &  2.26 & 2.02 & \cellcolor{gray!20} 52.50 \\

\noalign{\vskip 0.25ex}\cdashline{1-6}\noalign{\vskip 0.75ex}

{\textbf{Self-Refine}}
&  2.09 & 0.44  &  2.17 & 2.04 & \cellcolor{gray!20} 52.13 \\

\multirowcell{1}[-0.0ex][l]{\textbf{SPP}}
& 1.94  & 0.41  & 1.96  & 1.93 & \cellcolor{gray!20} 45.63 \\

\noalign{\vskip 0.25ex}\cdashline{1-6}\noalign{\vskip 0.75ex}

{\textbf{LLMvLLM}} 
& 1.28  & 0.25  &  1.21 &1.72  & \cellcolor{gray!20} 21.28 \\

{\textbf{MADR}}
&  1.91 &  0.33 &  1.90 & 1.80 & \cellcolor{gray!20} 40.75 \\

{\textbf{MultiDebate}} 
&  1.90 &  0.43 &  1.83 & 1.93 & \cellcolor{gray!20} 44.00 \\

\noalign{\vskip 0.25ex}\cdashline{1-6}\noalign{\vskip 0.75ex}

\rowcolor{blue!5} {\textbf{MARA (Ours)}}
& \textbf{2.31}  & \textbf{0.44}  &  \textbf{2.37} & \textbf{2.52} & \cellcolor{gray!20} \textbf{63.00} \\

\bottomrule
\end{tabular}
}
\vspace{-0.025in}
\end{table}

\begin{table}[t!]
\caption{\small Results on the Ubuntu Dialogue Corpus, using Claude as the base LLM. Best results highlighted in \textbf{bold}.}
\vspace{-0.1in}
\label{tab:ubuntu}
\small
\centering
\resizebox{0.475\textwidth}{!}{
\renewcommand{\arraystretch}{1.0}
\begin{tabular}{l ccccc}
\toprule
\textbf{Methods} & \textbf{Coh.} & \textbf{Grd.} & \textbf{Nat.} & \textbf{Eng.} & \textbf{Overall}\\

\midrule
\midrule

{\textbf{No Refine}}
& 2.44  & 0.62 & 2.09 & 1.84 & \cellcolor{gray!20} 57.57 \\

\noalign{\vskip 0.25ex}\cdashline{1-6}\noalign{\vskip 0.75ex}

{\textbf{Self-Refine}}
&  2.13 & 0.44 &  1.97 &  1.73 & \cellcolor{gray!20} 46.38 \\

\multirowcell{1}[-0.0ex][l]{\textbf{SPP}}
& 2.25  & 0.54 & 1.97 & 1.82 & \cellcolor{gray!20} 51.50 \\

\noalign{\vskip 0.25ex}\cdashline{1-6}\noalign{\vskip 0.75ex}

{\textbf{LLMvLLM}} 
& 1.48  & 0.24 & 1.28 & 1.38 & \cellcolor{gray!20} 20.33 \\

{\textbf{MADR}}
&  2.09 & 0.53 & 1.77 & 1.66 & \cellcolor{gray!20} 44.78 \\

{\textbf{MultiDebate}} 
& 1.98  & 0.48 & 1.70 & 1.67 & \cellcolor{gray!20} 41.38 \\

\noalign{\vskip 0.25ex}\cdashline{1-6}\noalign{\vskip 0.75ex}

\rowcolor{blue!5} {\textbf{MARA (Ours)}}
& \textbf{2.53}  & \textbf{0.70} & \textbf{2.25} & \textbf{2.33} & \cellcolor{gray!20} \textbf{68.88} \\ 

\bottomrule
\end{tabular}
}
\vspace{-0.075in}
\end{table}

\vspace{0.075in}
\noindent \textbf{Evaluation on the Role-Playing Dataset.}
While both role-playing conversations and our conversational setting involve persona-driven responses, their objectives are fundamentally different. Specifically, role-playing tasks focus on the LLM's persona, requiring an LLM to adopt a predefined persona, whereas ours focuses on the user's persona, enabling an LLM to generate responses that align with the user’s persona. Nonetheless, to further evaluate the effectiveness of MARA, we evaluate the performance on the role-playing conversational dataset, PRODIGy~\citep{PRODIGy}. As shown in Table~\ref{tab:roleplay}, MARA significantly outperforms other baselines, demonstrating its effectiveness even in role-playing conversational settings.

\vspace{0.075in}
\noindent \textbf{Evaluation on a Domain-Specific Corpus.}
To examine whether MARA generalizes well to domain-specific settings, we additionally evaluate it on the Ubuntu Dialogue Corpus~\citep{ubuntu}, consisting of multi-turn dialogues focused on technical problem-solving in software environments. As shown in Table~\ref{tab:ubuntu}, MARA outperforms all baselines, demonstrating its effectiveness even in software engineering–related conversational tasks.

\begin{table}[t!]
\caption{\small Results from human evaluation and Spearman's correlation with G-Eval on the same subset of the FoCus dataset.}
\vspace{-0.1in}
\label{tab:human_eval}
\small
\centering
\resizebox{0.475\textwidth}{!}{
\begin{tabular}{lccccc}
\toprule
\textbf{Methods} & \textbf{Coh.} & \textbf{Grd.} & \textbf{Nat.} & \textbf{Eng.} & \textbf{Overall} \\
\midrule\midrule
\textbf{No Refine} & 2.53 & 0.57 & 2.47 & 2.10 & 65.50 \\
\textbf{Self-Refine} & 2.04 & 0.36 & 1.86 & 1.93 & 44.38 \\
\textbf{SPP} & 2.49 & 0.53 & 2.44 & 2.18 & 64.63 \\
\textbf{MADR} & 2.03 & 0.35 & 2.10 & 1.65 & 43.50 \\
\textbf{MARA (Ours)} & \textbf{2.69} & \textbf{0.79} & \textbf{2.61} & \textbf{2.75} & \textbf{82.88} \\
\noalign{\vskip 0.25ex}\cdashline{1-6}\noalign{\vskip 0.75ex}
\textbf{Spearman's $\rho$} & 0.51 & 0.48 & 0.35 & 0.58 & -- \\
\bottomrule
\end{tabular}
}
\vspace{-0.025in}
\end{table}

\begin{table}[t!]
\caption{\small Ablation studies on each refining agent in FoCus, including incorporation of all perspectives into a single agent.}
\vspace{-0.1in}
\label{tab:refiningAgent_ablation}
\small
\centering
\resizebox{0.475\textwidth}{!}{
\renewcommand{\tabcolsep}{1.2mm}
\begin{tabular}{lcccccc}
\toprule
\textbf{Refining Agent} & \textbf{Coh.} & \textbf{Grd.} & \textbf{Nat.} & \textbf{Eng.} & \textbf{Overall} \\

\midrule
\midrule

\multirowcell{1}[-0.0ex][l]{\textbf{No Refine}}
& 2.39 & 0.49 & 2.02 & 2.15 & 56.75 \\

\noalign{\vskip 0.25ex}\cdashline{1-6}\noalign{\vskip 0.75ex}

\multirowcell{1}[-0.0ex][l]{\textbf{w/ Fact}}
& 2.62 & 0.61 & 2.14 & 2.52 & 68.75 \\

\multirowcell{1}[-0.0ex][l]{\textbf{w/ Persona}} 
& 2.54 & 0.49 & \textbf{2.18} & 2.77 & 68.38 \\

\multirowcell{1}[-0.0ex][l]{\textbf{w/ Coherence}} 
& 2.46 & 0.53 & 1.89 & 2.57 & 62.25 \\

\noalign{\vskip 0.25ex}\cdashline{1-6}\noalign{\vskip 0.75ex}

\multirowcell{1}[-0.0ex][l]{\textbf{w/ Single}}
& 2.54 & 0.58 & 2.15 & 2.55 & 67.50 \\

\multirowcell{1}[-0.0ex][l]{\textbf{w/ Single + Iteration}}
& 2.38 & 0.50 & 2.13 & 2.39 & 61.20 \\

\noalign{\vskip 0.25ex}\cdashline{1-6}\noalign{\vskip 0.75ex}

\multirowcell{1}[-0.0ex][l]{\textbf{MARA (Ours)}}
& \textbf{2.67} & \textbf{0.65} & 2.15 & \textbf{2.83} & \textbf{74.38}\\

\bottomrule
\end{tabular}
}
\vspace{-0.075in}
\end{table}

\subsection{Ablations and Discussion}
In this section, we provide a detailed analysis of the performance improvements.

\vspace{0.075in}
\noindent \textbf{Human Evaluation.}
Although G-Eval is a widely used evaluation metric for its high correlation with humans, we further conducted a human evaluation to measure the alignment between their evaluations. Specifically, 8 English-fluent participants were involved in evaluating 288 conversational queries from the FoCus dataset. As shown in Table~\ref{tab:human_eval}, human raters consistently preferred MARA’s responses over baseline models.

Additionally, to validate the reliability of G-Eval, we further assess the alignment between G-Eval and human judgments by calculating Spearman's rank correlation between the model-based evaluation and human evaluation results. The correlation values for coherence (0.51) and groundedness (0.48) indicate a reasonable alignment between G-Eval and human rankings, suggesting that G-Eval is fairly reliable in assessing these metrics. For engagingness, which reflects how the response aligns with the user's persona, the correlation of 0.58 demonstrates the strongest alignment, indicating that G-Eval is particularly effective at evaluating how engaging a conversation is. However, the relatively low correlation for naturalness (0.35) reveals a noticeable gap between G-Eval's assessments and human evaluations, especially in capturing the human-like qualities of the responses. This result might help explain the relatively smaller gaps in naturalness among the models, as shown in Table~\ref{tab:main}. 

\vspace{0.075in}
\noindent \textbf{Ablation Studies on Refining Agents.}
In order to see how each refining agent contributes to the overall performance, we perform ablation studies when adding three refining agents. As shown in Table~\ref{tab:refiningAgent_ablation}, each agent plays a crucial role in improving overall performance, and incorporating all three perspectives turns out to be even more beneficial. We also evaluate a single-agent variant that integrates all three aspects, which performs better than the No Refine baseline, validating the effectiveness of our proposed perspectives. However, MARA achieves superior performance, demonstrating the benefit of distributed specialization across multiple agents. Additionally, we examine an iterative single-agent setup, where the same agent refines responses across multiple rounds. This approach results in a performance drop, suggesting that initial errors may be amplified through repeated iterations—further underscoring the advantage of collaborative multi-agent refinement.

\begin{table}[t!]
\caption{\small Results with simultaneous (Sim.) and sequential (Seq.) communication strategies on the FoCus dataset.}
\vspace{-0.1in}
\label{tab:communication_ablation}
\small
\centering
\resizebox{0.475\textwidth}{!}{
\begin{tabular}{lcccccc}
\toprule
\textbf{Strategy} & \textbf{Coh.} & \textbf{Grd.} & \textbf{Nat.} & \textbf{Eng.} & \textbf{\#Agent} \\

\midrule
\midrule

\multirowcell{1}[-0.0ex][l]{\textbf{Sim. (F$+$C$+$P)}}
& 2.57 & 0.59 & 1.93 & 2.75 & 5.0 \\

\noalign{\vskip 0.25ex}\cdashline{1-6}\noalign{\vskip 0.75ex}

\multirowcell{1}[-0.0ex][l]{\textbf{Seq. (F$\rightarrow$C$\rightarrow$P)}}
& 2.54 & 0.56 & 1.99  & \textbf{2.85} & 4.0 \\

\multirowcell{1}[-0.0ex][l]{\textbf{Seq. (C$\rightarrow$P$\rightarrow$F)}} 
& 2.65 & 0.60 &  2.15 & 2.66 & 4.0 \\

\multirowcell{1}[-0.0ex][l]{\textbf{Seq. (P$\rightarrow$F$\rightarrow$C)}}
& 2.56 & 0.59 & 1.96  & 2.68 & 4.0 \\

\noalign{\vskip 0.25ex}\cdashline{1-6}\noalign{\vskip 0.75ex}

\multirowcell{1}[-0.0ex][l]{\textbf{MARA (Ours)}}
& \textbf{2.67} & \textbf{0.65} & \textbf{2.15} & 2.83 & 4.4 \\

\bottomrule
\end{tabular}
}
\vspace{-0.075in}
\end{table}

\vspace{0.075in}
\noindent \textbf{Ablation Studies on Communication Strategy.}
In order to investigate the efficacy of our proposed dynamic sequential communication strategy, we compare it against other possible strategies using our refining agents. As shown in Table~\ref{tab:communication_ablation}, the simultaneous communication strategy requires more access to the agents per query, as it involves all three refining agents as well as the additional finalizer agent. We also evaluate the performance of three different sequential settings, where the agents refine the initial response in static orders. The results indicate that performance varies significantly depending on the refinement order, making it impractical for realistic conversational settings, as users would need to predefine the optimal sequence for every query. On the other hand, our dynamic strategy adapts to each query by selecting the most relevant sequence of agents, leading to more consistent and improved performance in realistic conversational settings. 

\vspace{0.075in}
\noindent \textbf{Analyses on Fact-Refining Agent.}
A notable advantage of the multi-agent framework is its flexibility in utilizing different agents, and in our case, we can assign a different model to the fact-refining agent. 
As shown in Table~\ref{tab:main} and Table~\ref{tab:factchecking}, while MARA with a fact-refining agent of the same model size as a responding agent indeed improves performance when compared to baselines, the results show that using a different fact-refining agent yields much better outcomes. This highlights a key strength of the multi-agent framework: the ability to utilize different LLMs for specific roles, thereby enhancing performance beyond what a single-agent approach can achieve. Then, one might ask why not simply use a more powerful model as the responding agent directly. However, Table~\ref{tab:factchecking} shows that refinement with MARA significantly improves performance, even when the responding agent is already powerful. This indicates that as LLMs continue to advance, adopting MARA is expected to further amplify their effectiveness.

\begin{table}[t!]
\caption{\small Analyses of the fact-refining agent size and the responding agent size in the FoCus dataset.}
\vspace{-0.1in}
\label{tab:factchecking}
\small
\centering
\resizebox{0.475\textwidth}{!}{
\renewcommand{\arraystretch}{1.0}
\begin{tabular}{lccccc}
\toprule
\textbf{Model Size} & \textbf{Coh.} & \textbf{Grd.} & \textbf{Nat.} & \textbf{Eng.} \\

\midrule
\midrule

\multirowcell{1}[-0.0ex][l]{\textbf{No Refine w/ Sonnet 3}} 
& 2.39 & 0.49 &  2.02  & 2.15  \\

\multirowcell{1}[-0.0ex][l]{\textbf{MARA w/ Fact Sonnet 3}}
& 2.56 & 0.57 & 2.00  & 2.80 \\

\multirowcell{1}[-0.0ex][l]{\textbf{MARA w/ Fact Sonnet 3.5}}
& \textbf{2.67} & \textbf{0.65} & \textbf{2.15} & \textbf{2.83} \\

\noalign{\vskip 0.25ex}\cdashline{1-5}\noalign{\vskip 0.75ex}

\multirowcell{1}[-0.0ex][l]{\textbf{No Refine w/ Sonnet 3.5}} 
& 2.50 & 0.58 & 2.10 & 2.32 \\

\multirowcell{1}[-0.0ex][l]{\textbf{MARA w/ Sonnet 3.5}} 
& \textbf{2.87} & \textbf{0.78} & \textbf{2.55} & \textbf{2.91} \\

\bottomrule
\end{tabular}
}
\end{table}

\begin{table}[t!]
\caption{\small GPT4o-mini and Llama 3.1 8B, 70B on FoCus.}
\vspace{-0.1in}
\label{tab:other_llm}
\small
\centering
\resizebox{0.475\textwidth}{!}{
\renewcommand{\arraystretch}{1.0}
\begin{tabular}{llcccc}
\toprule
& \textbf{Methods} & \textbf{Coh.} & \textbf{Grd.} & \textbf{Nat.} & \textbf{Eng.} \\
\midrule
\midrule
\multirow{2}{*}{\rotatebox[origin=c]{90}{\scriptsize{\textbf{GPT}}}} 
 & \textbf{No Refine} & 2.45 & 0.47 & 2.27 & 1.98 \\
 & \textbf{MARA (Ours)} & \textbf{2.61} & \textbf{0.57} & \textbf{2.32} & \textbf{2.70}\\
 
\cdashline{1-6}\noalign{\vskip 0.75ex}

\multirow{4}{*}{\rotatebox[origin=c]{90}{\scriptsize{
 \textbf{Llama}}}}

 & \textbf{No Refine - 8B} & 2.30 & 0.39 & 2.09 & 1.88 \\
 & \textbf{MARA - 8B (Ours)} & \textbf{2.46} & \textbf{0.48} & \textbf{2.17} & \textbf{2.24} \\

\cdashline{2-6}\noalign{\vskip 0.75ex}

 & \textbf{No Refine - 70B} & 2.41 & 0.44 & 2.25 & 1.88 \\
 & \textbf{MARA - 70B (Ours)} & \textbf{2.60} & \textbf{0.55} & \textbf{2.31} & \textbf{2.67} \\

\bottomrule
\end{tabular}
}
\vspace{-0.075in}
\end{table}

\vspace{0.075in}
\noindent \textbf{Effectiveness on Diverse LLMs.}
We further validate the effectiveness of MARA with other widely used LLMs in Table~\ref{tab:other_llm}. The results demonstrate that MARA can effectively refine initial responses across different LLMs, seamlessly integrate with diverse models, and enhance their outputs, highlighting its broad applicability to various LLMs.

\vspace{0.075in}
\noindent \textbf{Case Studies.}
Lastly, we present case studies in Table~\ref{tab:case_study} and an illustrative example showing the roles of refining agents and the planner in Table~\ref{tab:example}, along with analyses in Appendix~\ref{appendix:results}.

\section{Conclusion}
While LLMs have shown promise in conversational systems, they often struggle with complex, customized conversations requiring persona alignment and specific knowledge. Prior approaches using single-agent self-refinement can be suboptimal, as one model must handle all aspects of feedback and refinement. In this work, we presented a cooperative multi-agent framework to address these limitations, with specialized agents for fact-checking, persona alignment, and coherence, by allocating them dynamically tailored to each query, and it significantly outperforms existing baselines. Our evaluation on challenging conversational datasets shows that this multi-agent framework significantly outperforms existing baselines.

\vspace{-0.025in}
\section*{Limitations}
\vspace{-0.05in}

While our MARA framework demonstrates clear advantages by adaptively selecting the most suitable refining agents for diverse customized conversational settings, there remains room for improvement, particularly in the planner agent, as shown by the performance gap between the ideal planner and the current planner Table~\ref{tab:planner}. Specifically, since our current planner is fully based on an unsupervised LLM, constructing a dataset with labeled ideal sets and sequences of required perspectives, followed by fine-tuning the planner on this data, could be a promising direction for future research. 

While a multi-agent framework shows remarkable effectiveness, it may raise concerns about scalability and resource efficiency. Therefore, we further suggest some possible approaches that can address these challenges. As demonstrated by the significantly reduced number of LLM accesses with an optimized planner in Table~\ref{tab:planner}, improving the planner agent can also alleviate concerns regarding resource efficiency. To further enhance scalability, exploring lightweight or more efficient agent models would be also beneficial. In Table~\ref{tab:factchecking} and Table~\ref{tab:other_llm}, we show that our proposed MARA performs well even with a smaller LLM. Notably, Table~\ref{tab:other_llm} shows significant improvements with a lightweight, open-sourced LLM, the Llama 3.1 8B model, which highlights its potential for scalability. Nonetheless, further exploration of scalability remains a promising avenue for future work, which we leave as a meaningful future work.

Also, the flexibility of the multi-agent framework in selecting diverse external tools or LLMs opens up opportunities for incorporating tools such as Retrieval-Augmented Generation (RAG) systems to further enhance performance, which could also be an interesting future research direction.
\section*{Ethics Statement}
With the goal of developing human-centric applications that assist users in conversations, we experimentally validate the effectiveness of MARA for its applicability in realistic conversational scenarios, where a diverse range of queries with varying requests exists. However, given the potential diversity of real-world user inputs, it is crucial to consider scenarios where inputs or generated responses may be offensive or harmful~\citep{DBLP:conf/acl/ShinSLJP24, llm_conv_survey1}.
To ensure safe and responsible deployment, additional safeguards are necessary. In particular, integrating an agent to monitor both user inputs and generated content for harmful or offensive material would be valuable. We consider this a critical area for future research, aiming to improve MARA’s robustness and ethical alignment in real-world applications.

\bibliography{custom}

\appendix

\clearpage

\appendix

\section{Additional Experimental Setups}
\label{appendix:experimental_setup}

\subsection{Datasets}
We validate MARA in a realistic conversational setting by testing it on three conversational datasets, including persona-aligned and knowledge-grounded queries.

\noindent \textbf{1) FoCus}~\citep{focus} consists of conversational queries requiring both persona-aligned and knowledge-grounded responses, with knowledge sourced from Wikipedia. The dataset has an average of 11.9 conversational turns. We use the publicly available validation set.

\noindent \textbf{2) PersonaChat}~\citep{personachat} is a conversational dataset involving dialogues between two speakers, each having persona sentences that describe their character and serve as context for the conversation. The dataset has an average of 14.0 conversational turns. We use the publicly open validation set.

\noindent \textbf{3) INSCIT}~\citep{inscit} is an information-seeking conversational dataset, where the knowledge is grounded in Wikipedia. The dataset has an average of 11.8 conversational turns. We use the publicly accessible test set for the evaluation.

\subsection{Models}
We compare MARA against four baseline models, starting with a baseline without refinement, two single-agent refinement approaches, and one multi-agent refinement approach.

\noindent \textbf{1) No Refine} is an approach involves generating a response solely with a responding agent, without any further refinement of the response.

\noindent \textbf{2) Self-Refine}~\citep{self-refine} is a representative single agent refinement baseline, which makes an LLM to generate feedback on 10 aspects (e.g., Relevant, Informative, Engaging) of its own response and iteratively refines the output by incorporating this feedback.

\noindent \textbf{3) SPP}~\citep{spp} is another single-agent refinement baseline where a single agent generates multiple self-generated personas and makes them collaborate to solve a task using a single prompt.

\noindent \textbf{4) LLMvLLM}~\citep{llmvllm} is a multi-agent refinement baseline where two agents engage in cross-examination to detect factual errors in generated response. One agent (EXAMINEE) produces an initial statement, while the other agent (EXAMINER) iteratively questions it, aiming to identify inconsistencies through multiple rounds of interaction.

\noindent \textbf{5) MADR}~\citep{madr} is a multi agent refinement baseline where two agents debate based on the predefined error types to generate faithful fact checking explanations. They iteratively debate until the judge agent decides that two agents are in consensus, and the finalizer agent refines the refined response based on the feedback generated during a debate.

\noindent \textbf{6) MultiDebate}~\citep{multidebate} is a multi-agent refinement baseline where multiple language model agents iteratively engage in a structured debate to improve factual accuracy and reasoning. Each agent independently generates an initial response to a given query, followed by multiple rounds of critique and refinement based on responses from other agents.

\noindent \textbf{7) MARA} is our proposed system, where a planner agent dynamically determines the optimal sequence of refining agents to involve, and then the agents refine the response in the specified order.

\subsection{Implementation Details}
The prompts used for MARA are provided in Table~\ref{tab:prompt:responding_agent} (responding agent), Table~\ref{tab:prompt:planner_agent} (planner agent), Table~\ref{tab:prompt:fact_refining_agent} (fact-refining agent), Table~\ref{tab:prompt:persona_refining_agent} (persona-refining agent), and Table~\ref{tab:prompt:coherence_refining_agent} (coherence-refining agent). The prompt templates for G-Eval are shown in Table~\ref{tab:prompt:geval_coherence}, Table~\ref{tab:prompt:geval_groundedness}, Table~\ref{tab:prompt:geval_naturalness}, and Table~\ref{tab:prompt:geval_engagingness}.

\section{Experimental Results}
\label{appendix:results}

\subsection{Significance Test on FoCus (Table~\ref{tab:significance_focus})}

\paragraph{Coherence}
The one-way ANOVA reveals a significant effect of model type on coherence scores ($F(3, 8) = 763.67, p < 0.001$), indicating that at least one model has a significantly different mean score. Post-hoc Tukey HSD tests show that MARA significantly outperforms all other models ($p < 0.001$), with a mean difference of $0.7456$ over the MADR model, $0.2838$ over the No Refine model, and $0.5691$ over the Self-Refine model. Additionally, the No Refine model outperforms both Self-Refine and SPP, with a small but significant difference of $0.0521$ over SPP. Self-Refine also outperforms SPP with a mean difference of $0.3375$. Overall, MARA is the top performer, offering substantial improvements in coherence, demonstrating that its enhancements contribute meaningfully to better response quality across all models.

\paragraph{Groundedness}
The one-way ANOVA reveals a significant effect of model type on groundedness scores ($F(3, 8) = 426.81, p < 0.001$), indicating notable differences in performance among the models. Post-hoc Tukey HSD tests show that MARA achieves higher groundedness scores than all other models ($p < 0.001$), with mean differences of $0.3337$ over MADR, $0.1648$ over No Refine, and $0.2806$ over Self-Refine. Additionally, No Refine surpasses Self-Refine by $0.1158$, while its difference from SPP is not significant. Self-Refine also shows an advantage over SPP, with a mean difference of $0.101$. These results highlight the effectiveness of MARA in improving groundedness compared to other approaches.

\paragraph{Naturalness}
The one-way ANOVA reveals a significant effect of model type on naturalness scores ($F(3, 8) = 215.79, p < 0.001$), indicating that model differences lead to varying levels of naturalness. Post-hoc Tukey HSD tests show that MARA significantly outperforms all other models ($p < 0.001$), with mean differences of $0.4772$ over MADR, $0.1339$ over No Refine, and $0.2851$ over Self-Refine. Furthermore, No Refine performs better than Self-Refine by $0.1511$, although its comparison with SPP shows no significant difference. Self-Refine outperforms SPP with a mean difference of $0.1235$. Overall, these results indicate that MARA demonstrates strong naturalness compared to other models, with clear advantages over the other approaches.

\paragraph{Engagingness}
The one-way ANOVA reveals a significant effect of model type on engagingness scores ($F(3, 8) = 1772.37, p < 0.001$), indicating notable differences across the models. Post-hoc Tukey HSD tests show that MARA significantly outperforms all other models ($p < 0.001$), with mean differences of $1.2939$ over MADR, $0.6771$ over No Refine, and $0.7801$ over Self-Refine. Furthermore, No Refine outperforms Self-Refine by $0.1030$, and SPP by $0.3065$. Self-Refine also outperforms SPP with a mean difference of $0.4095$. Overall, these results demonstrate that MARA consistently achieves higher engagingness scores compared to the other models, highlighting its effectiveness in enhancing the engagingness of responses.

\subsection{Significance Test on PersonaChat (Table~\ref{tab:significance_personachat})}
\paragraph{Coherence}
The one-way ANOVA reveals a significant effect of model type on coherence scores ($F(3, 8) = 725.87, p < 0.001$), indicating substantial differences between the models. Post-hoc Tukey HSD tests show that MARA significantly outperforms all other models ($p < 0.001$), with mean differences of $0.7359$ over MADR, $0.4572$ over No Refine, and $0.0611$ over Self-Refine. Additionally, No Refine outperforms Self-Refine by $0.3962$ and SPP by $0.1141$. Self-Refine also outperforms SPP with a mean difference of $0.2821$. These results demonstrate that MARA leads to higher coherence scores, providing a clear advantage over the other models in this aspect.

\paragraph{Groundedness}
The one-way ANOVA reveals a significant effect of model type on groundedness scores ($F(3, 8) = 498.84, p < 0.001$), indicating notable differences between the models. Post-hoc Tukey HSD tests show that MARA significantly outperforms all other models ($p < 0.001$), with mean differences of $0.2731$ over MADR, $0.1406$ over No Refine, and $0.0369$ over Self-Refine. Furthermore, No Refine outperforms Self-Refine by $0.1037$, although its comparison with SPP shows no significant difference. Self-Refine also outperforms SPP with a mean difference of $0.0822$. These results indicate that MARA provides consistent improvements in groundedness compared to other models.

\paragraph{Naturalness}
The one-way ANOVA reveals a significant effect of model type on naturalness scores ($F(3, 8) = 641.13, p < 0.001$), indicating substantial differences across the models. Post-hoc Tukey HSD tests show that MARA significantly outperforms all other models ($p < 0.001$), with mean differences of $0.6000$ over MADR, $0.4273$ over No Refine, and $0.2599$ over SPP. The comparison between MARA and Self-Refine also shows a small but significant difference of $0.0896$. Furthermore, No Refine outperforms Self-Refine by $0.5169$ and SPP by $0.1673$, while Self-Refine significantly outperforms SPP by $0.3496$. These results highlight MARA’s strong performance in naturalness compared to other models.

\paragraph{Engagingness}
The one-way ANOVA reveals a significant effect of model type on engagingness scores ($F(3, 8) = 1028.47, p < 0.001$), showing that the models differ significantly in their performance. Post-hoc Tukey HSD tests indicate that MARA significantly outperforms all other models ($p < 0.001$), with mean differences of $1.2213$ over MADR, $0.9953$ over No Refine, and $0.4745$ over SPP. Additionally, No Refine outperforms Self-Refine by $0.7535$ and SPP by $0.5208$, while Self-Refine outperforms SPP by $0.2327$. These results underscore MARA's effectiveness in enhancing engagingness across different models.

\subsection{Significance Test on INSCIT (Table~\ref{tab:significance_inscit})}
\paragraph{Coherence}
The one-way ANOVA reveals a significant effect of model type on coherence scores ($F(3, 8) = 200.85, p < 0.001$), indicating that the models perform differently. Post-hoc Tukey HSD tests show that MARA significantly outperforms MADR, with a mean difference of $0.3264$, as well as No Refine ($0.0844$) and SPP ($0.0946$). The comparison between MADR and Self-Refine is not statistically significant, but No Refine outperforms Self-Refine by $0.225$. Self-Refine also outperforms SPP with a mean difference of $0.2148$. Overall, MARA demonstrates stronger coherence performance compared to most other models.

\paragraph{Groundedness}
The one-way ANOVA reveals a significant effect of model type on groundedness scores ($F(3, 8) = 119.45, p < 0.001$), indicating that the models show distinct performance differences. Post-hoc Tukey HSD tests show that MARA significantly outperforms MADR with a mean difference of $0.129$, as well as No Refine ($0.0765$) and SPP ($0.101$). No Refine also outperforms Self-Refine by $0.0721$. However, the comparison between MADR and Self-Refine is not statistically significant. These results suggest that MARA provides meaningful improvements in groundedness compared to the other models.

\paragraph{Naturalness}
The one-way ANOVA reveals a significant effect of model type on naturalness scores ($F(3, 8) = 39.70, p < 0.001$), indicating differences in performance across models. Post-hoc Tukey HSD tests show that MARA significantly outperforms MADR by $0.1382$ and No Refine by $0.1052$, though the comparison between MARA and Self-Refine is not significant. MADR also outperforms No Refine by $0.2434$ and Self-Refine by $0.0707$, while No Refine outperforms Self-Refine by $0.1727$. These results demonstrate that MARA provides strong naturalness performance, with notable differences in certain comparisons, but not against all models.

\paragraph{Engagingness}
The one-way ANOVA reveals a significant effect of model type on engagingness scores ($F(3, 8) = 255.26, p < 0.001$), showing differences in model performance. Post-hoc Tukey HSD tests show that MARA significantly outperforms MADR by $0.8387$, No Refine by $0.4544$, and SPP by $0.2664$. The comparison between No Refine and Self-Refine is not statistically significant, but No Refine outperforms SPP by $0.188$, and Self-Refine outperforms SPP by $0.2253$. These results suggest that MARA provides stronger engagingness performance compared to the other models, with significant improvements over most.

\subsection{Ablation Studies on Design Choices}

To further investigate the effectiveness of our strategy, we conduct ablation studies on specific design choices. First, we examine the impact of allowing each refining agent to recognize the presence of other agents by passing the planner agent's output, which includes both the sequence and justification for the selected set and sequence. Specifically, the planner agent generates a suitable sequence with justification and passes it to the refining agents, enabling each agent to be aware of the preceding and following agents, as well as the rationale behind the sequence. As shown in Table~\ref{tab:verify_justify_ablation}, when the planner agent's output is not passed to the refining agents, performances with all metrics decrease, particularly in groundedness. This underscores the importance of enabling agents to be aware of one another to perform their roles and collaborate more effectively.

\begin{table}[t!]
\caption{\small Additional ablation studies in the FoCus dataset.}
\label{tab:verify_justify_ablation}
\small
\centering
\resizebox{0.475\textwidth}{!}{
\renewcommand{\arraystretch}{0.975}

\begin{tabular}{lccccc}
\toprule
\textbf{} & \textbf{Coh.} & \textbf{Grd.} & \textbf{Nat.} & \textbf{Eng.} \\

\midrule
\midrule

\multirowcell{1}[-0.0ex][l]{\textbf{MARA (Ours)}} 
& \textbf{2.67} & \textbf{0.65}  &  \textbf{2.15} & \textbf{2.83} \\

\noalign{\vskip 0.25ex}\cdashline{1-6}\noalign{\vskip 0.75ex}

\multirowcell{1}[-0.0ex][l]{\textbf{w/o Planner Outputs}}
&  2.62 & 0.60  & 2.12 & 2.75 \\

\multirowcell{1}[-0.0ex][l]{\textbf{w/o Step-by-Step}}
& 2.59 & 0.65  & 2.00  & 2.66 \\

\noalign{\vskip 0.25ex}\cdashline{1-6}\noalign{\vskip 0.75ex}

\multirowcell{1}[-0.0ex][l]{\textbf{No Refine}} 
&  2.39 & 0.49 & 2.02 & 2.15 \\

\bottomrule
\end{tabular}
}

\end{table}

Additionally, we explore whether a step-by-step for each refining agent's reasoning process—first verifying, then refining responses with justifications—is effective. As shown in Table~\ref{tab:verify_justify_ablation}, performance in coherence, naturalness, and engagingness decrease when this process is removed. This suggests that the verification step and justification is crucial for maintaining high-quality responses, as it allows for a more thorough assessment before refining.

\begin{table*}[t!]
\caption{\small Case study on the FoCus dataset comparing MARA with SPP.}
\vspace{-0.1in}
\label{tab:case_study}
\centering
\resizebox{\textwidth}{!}{
\renewcommand{\arraystretch}{1.0}
\begin{tabular}{ccccc}
\toprule

\multicolumn{1}{p{.2\textwidth}}{\textbf{Question}} & \multicolumn{1}{p{.2\textwidth}}{\textbf{Persona}} & \multicolumn{1}{p{.2\textwidth}}{\textbf{Fact}} & \multicolumn{1}{p{.4\textwidth}}{\textbf{SPP}} & \multicolumn{1}{p{.4\textwidth}}{\textbf{MARA (Ours)}} \\

\midrule \midrule

\multicolumn{1}{p{.2\textwidth}}{\textbf{Q1:} What is the height of the mountain?}&
\multicolumn{1}{p{.2\textwidth}}{I have the fantasy about valley. I like lakes. I would like to go to California. I have seen lava domes. I have heard about Lassen Volcanic National Park.} & \multicolumn{1}{p{.2\textwidth}}{Located 2 miles (3.2 km) to the north of Lassen Peak, they have an elevation of about 8,448 feet (2,575 m)} &
\multicolumn{1}{p{.4\textwidth}}{Unfortunately, without any additional context about what fictional universe, book, movie or game Chaos Crags is from, \textcolor{red}{it's very difficult for me to estimate the height of this mountain.} If you could provide some more details about the setting, terrain, or role this mountain plays in the story, that would help me make an educated guess about how tall Chaos Crags is likely intended to be. I'd be happy to take another look once I have some contextual clues from the source material.} &
\multicolumn{1}{p{.4\textwidth}}{The Chaos Crags are not a single mountain, but rather a group of six steep-sided lava domes or volcanic plugs. They rise approximately 550 meters (about 1,800 feet) above the Chaos Jumbles, which is the rockfall avalanche deposit at their base. 
To provide more context, the highest point of the Chaos Crags reaches an elevation of about \textcolor{blue} {2,575 meters (8,450 feet) above sea level.}}  \\ 

\noalign{\vskip 0.25ex}\cdashline{1-5}\noalign{\vskip 0.75ex}
\multicolumn{1}{p{.2\textwidth}}{\textbf{Q2: } Is there any lake near by?}&
\multicolumn{1}{p{.2\textwidth}}{I have the fantasy about valley. I like lakes. I would like to go to California. I have seen lava domes. I have heard about Lassen Volcanic National Park.} & \multicolumn{1}{p{.2\textwidth}}{Each year, a lake forms at the base of the Crags, and typically dries by the end of the summer season.} &
\multicolumn{1}{p{.4\textwidth}}{Based on the lore provided, Chaos Crags is an immense, treacherous mountain range in the northern reaches of Middle-earth. While the exact height is not specified, it is described as one of the tallest peaks in that region, with the highest spires potentially reaching elevations of 20,000-25,000 feet.
The Crags likely consist of jagged, ice-capped pinnacles towering miles into the sky above deep glacial valleys. Their sheer cliffs and avalanche-prone slopes would make them extremely hazardous and virtually unscalable.
\textcolor{red}{While no major lakes are referenced}, the Crags overlook the icy waters of the Bay of Forochel and its inlets. Nestled among the rocky chaos may be smaller glacier-fed tarns and streams cascading from the ice fields high above.
The overall scenery is one of bleak, desolate beauty - an otherworldly alpine expanse exposed to the punishing winds blowing off the northern seas.} &
\multicolumn{1}{p{.4\textwidth}}{\textcolor{blue} {As someone drawn to the beauty of lakes and the raw power of volcanic landscapes}, you'll be delighted to know there are several lakes nestled right near the incredible Chaos Crags formation in Lassen Volcanic National Park.
The most picturesque is undoubtedly Manzanita Lake, formed around 1,100 years ago by lava flows that dammed a stream during the same eruptive period that created the crags themselves. \textcolor{blue} {Can you imagine hiking along its shores, taking in views of those dramatic lava domes rising in the backdrop?} A scene like that perfectly encapsulates the awe-inspiring volcanic wonders of this region.
But Manzanita isn't the only lake beckoning explorers like yourself. Reflection Lake and Chaos Crags Lake, though smaller, offer equally enticing opportunities to immerse yourself in this geologically active landscape.}  \\ 

\bottomrule

\end{tabular}
}

\end{table*}
\begin{table*}[t!]
\caption{Example illustrating the roles of refining agents and the planner.}
\vspace{-0.1in}
\label{tab:example}
\scriptsize
\centering
\resizebox{\textwidth}{!}{
\renewcommand{\arraystretch}{1.1}
\begin{tabular}{l c}
\toprule
\textbf{Question} & 
\multicolumn{1}{p{.9\textwidth}}{
Wow, this is amazing! What is this?} \\
\noalign{\vskip 0.25ex}\cdashline{1-2}\noalign{\vskip 0.75ex}
\textbf{Initial Response} & 
\multicolumn{1}{p{.9\textwidth}}{
I apologize, but I don't have any specific information about what you're referring to when you say "this." Without more context or details about what you're looking at or experiencing, I can't provide an accurate answer. If you could provide more information or describe what you're seeing, I'd be happy to try and help explain it to you.
} \\
\noalign{\vskip 0.25ex}\cdashline{1-2}\noalign{\vskip 0.75ex}
\textbf{Planner Decision} & 
\multicolumn{1}{p{.9\textwidth}}{
Selected agents: Coherence, Persona (Fact not needed)
} \\
\noalign{\vskip 0.25ex}\cdashline{1-2}\noalign{\vskip 0.75ex}
\textbf{Justification of Planner Decision} & 
\multicolumn{1}{p{.9\textwidth}}{
1. Coherence: This agent should go first to address the lack of context in the initial response. 2. Persona: After improving coherence, this agent can tailor the response to align with the user's interests in history, particularly related to Vermont and historic houses. The Fact Agent is not necessary in this case, as there are no specific facts to verify in the initial response.
} \\
\noalign{\vskip 0.25ex}\cdashline{1-2}\noalign{\vskip 0.75ex}
\textbf{Refined Response (Coherence)} & 
\multicolumn{1}{p{.9\textwidth}}{
Ah, I see you've discovered the Galusha House! It's truly a remarkable piece of history. This historic house is indeed something to be excited about. It's a well-preserved example of early American architecture and has a fascinating story behind it.
} \\
\noalign{\vskip 0.25ex}\cdashline{1-2}\noalign{\vskip 0.75ex}
\textbf{Refined Response (Persona)} & 
\multicolumn{1}{p{.9\textwidth}}{
You've stumbled upon a true gem of Vermont's history - the Galusha House! As someone who appreciates historic houses, you're in for a treat. This well-preserved example of early architecture isn't just beautiful; it's also steeped in fascinating stories, including ties to the Underground Railroad. Would you like to learn more about its role in history or its architectural features?
} \\

\bottomrule

\end{tabular}
}

\end{table*}

\subsection{Case Studies}
Lastly, we present case studies in Table~\ref{tab:case_study}, which shows two consecutive turns within a conversation about the `Chaos Crags'. In the first turn, SPP fails to provide an accurate response, losing track of the conversation's context and offering irrelevant information about fictional mountains instead of directly addressing the user's query about the mountain's height. In contrast, our MARA model delivers an accurate and contextually appropriate response, correctly identifying Chaos Crags and providing the exact elevation, enriched with additional context about its geological formation. This demonstrates MARA's ability to maintain focus and generate factually grounded responses. 

In the second turn, both SPP and MARA attempt to understand the context of the user's question. However, SPP not only fails to directly address the user's question about a nearby lake, stating, `While no major lakes are referenced,' but also lacks engagement with the user. MARA, on the other hand, not only provides the correct factual information but also tailors the response to the user's expressed interest in lakes and volcanic landscapes, creating a more personalized and engaging interaction. Furthermore, MARA takes the conversation a step further by asking a follow-up question: `Can you imagine hiking along its shores, taking in views of those dramatic lava domes rising in the backdrop?'. This question invites the user to visualize the experience, fostering deeper engagement. Such interactions showcase MARA’s strength in incorporating user preferences beyond fact delivery.

\subsection{Illustrative Example of Agent Roles}
While we included a case study example in Table~\ref{tab:case_study}, we introduce another example to provide a more detailed description of the agents in Table~\ref{tab:example}. In this example, after reviewing the query and initial response, the planner selects the coherence-refining agent and persona-refining agent as the required agents. Following the planner’s decision, the coherence-refining agent first addresses the initial response’s lack of context and introduces the Galusha House, while the persona-refining agent further refines the response by adding personalized context and engaging follow-up.

\newpage

\begin{table*}[t!]
\caption{\small Significance Testing for Coherence, Groundedness, Naturalness, and Engagingness (FoCus Dataset)}
\vspace{-0.1in}
\label{tab:significance_focus}
\small
\centering
\resizebox{1.0\textwidth}{!}{
\renewcommand{\arraystretch}{0.975}
\begin{tabular}{lcccccccc}
\toprule
& \textbf{Comparison} & \textbf{Model 1} & \textbf{Model 2} & \textbf{Mean Diff.} & \textbf{P-adj} & \textbf{Lower} & \textbf{Upper} & \textbf{Significant?} \\
\midrule
\midrule
\multirowcell{8}[-0.0ex][c]{\rotatebox[origin=c]{90}{\textbf{Coherence}}} 
& \multirowcell{1}[-0.0ex][l]{\textbf{No Refine vs MARA (Ours)}} & No Refine & MARA (Ours) & -0.2838 & 0.000 & -0.3332 & -0.2344 & Yes \\

& \multirowcell{1}[-0.0ex][l]{\textbf{No Refine vs MADR}} 
& No Refine & MADR & -0.4619 & 0.000 & -0.5153 & -0.4084 & Yes \\

& \multirowcell{1}[-0.0ex][l]{\textbf{No Refine vs Self-Refine}} 
& No Refine & Self-Refine & -0.2854 & 0.000 & -0.3348 & -0.2360 & Yes \\

& \multirowcell{1}[-0.0ex][l]{\textbf{No Refine vs SPP}} 
& No Refine & SPP & 0.0521 & 0.0378 & 0.0027 & 0.1015 & Yes \\

& \multirowcell{1}[-0.0ex][l]{\textbf{MARA (Ours) vs MADR}} 
& MARA (Ours) & MADR & 0.7456 & 0.000 & 0.6962 & 0.7950 & Yes \\

& \multirowcell{1}[-0.0ex][l]{\textbf{MARA (Ours) vs Self-Refine}} 
& MARA (Ours) & Self-Refine & 0.5691 & 0.000 & 0.5197 & 0.6185 & Yes \\

& \multirowcell{1}[-0.0ex][l]{\textbf{MARA (Ours) vs SPP}} 
& MARA (Ours) & SPP & -0.2317 & 0.000 & -0.2811 & -0.1823 & Yes \\

& \multirowcell{1}[-0.0ex][l]{\textbf{Self-Refine vs SPP}} 
& Self-Refine & SPP & 0.3375 & 0.000 & 0.2881 & 0.3869 & Yes \\

\midrule
\midrule

\multirowcell{8}[-0.0ex][c]{\rotatebox[origin=c]{90}{\textbf{Groundedness}}}
& \multirowcell{1}[-0.0ex][l]{\textbf{No Refine vs MARA (Ours)}} 
& No Refine & MARA (Ours) & -0.1648 & 0.000 & -0.1937 & -0.1359 & Yes \\

& \multirowcell{1}[-0.0ex][l]{\textbf{No Refine vs MADR}} 
& No Refine & MADR & 0.1689 & 0.000 & 0.1400 & 0.1978 & Yes \\

& \multirowcell{1}[-0.0ex][l]{\textbf{No Refine vs Self-Refine}} 
& No Refine & Self-Refine & -0.1158 & 0.000 & -0.1447 & -0.0869 & Yes \\

& \multirowcell{1}[-0.0ex][l]{\textbf{No Refine vs SPP}} 
& No Refine & SPP & -0.0148 & 0.4793 & -0.0437 & 0.0140 & No \\

& \multirowcell{1}[-0.0ex][l]{\textbf{MARA (Ours) vs MADR}} 
& MARA (Ours) & MADR & 0.3337 & 0.000 & 0.3048 & 0.3626 & Yes \\

& \multirowcell{1}[-0.0ex][l]{\textbf{MARA (Ours) vs Self-Refine}} 
& MARA (Ours) & Self-Refine & 0.2806 & 0.000 & 0.2517 & 0.3095 & Yes \\

& \multirowcell{1}[-0.0ex][l]{\textbf{MARA (Ours) vs SPP}} 
& MARA (Ours) & SPP & -0.1796 & 0.000 & -0.2085 & -0.1508 & Yes \\

& \multirowcell{1}[-0.0ex][l]{\textbf{Self-Refine vs SPP}} 
& Self-Refine & SPP & 0.1010 & 0.000 & 0.0721 & 0.1298 & Yes \\

\midrule
\midrule

\multirowcell{8}[-0.0ex][c]{\rotatebox[origin=c]{90}{\textbf{Naturalness}}}
& \multirowcell{1}[-0.0ex][l]{\textbf{No Refine vs MARA (Ours)}} 
& No Refine & MARA (Ours) & -0.1339 & 0.0001 & -0.1909 & -0.0770 & Yes \\

& \multirowcell{1}[-0.0ex][l]{\textbf{No Refine vs MADR}} 
& No Refine & MADR & 0.3432 & 0.000 & 0.2863 & 0.4002 & Yes \\

& \multirowcell{1}[-0.0ex][l]{\textbf{No Refine vs Self-Refine}} 
& No Refine & Self-Refine & -0.1511 & 0.000 & -0.2081 & -0.0942 & Yes \\

& \multirowcell{1}[-0.0ex][l]{\textbf{No Refine vs SPP}} 
& No Refine & SPP & -0.0277 & 0.5296 & -0.0846 & 0.0293 & No \\

& \multirowcell{1}[-0.0ex][l]{\textbf{MARA (Ours) vs MADR}} 
& MARA (Ours) & MADR & 0.4772 & 0.000 & 0.4202 & 0.5341 & Yes \\

& \multirowcell{1}[-0.0ex][l]{\textbf{MARA (Ours) vs Self-Refine}} 
& MARA (Ours) & Self-Refine & 0.2851 & 0.000 & 0.2281 & 0.3420 & Yes \\

& \multirowcell{1}[-0.0ex][l]{\textbf{MARA (Ours) vs SPP}} 
& MARA (Ours) & SPP & -0.1616 & 0.000 & -0.2185 & -0.1046 & Yes \\

& \multirowcell{1}[-0.0ex][l]{\textbf{Self-Refine vs SPP}} 
& Self-Refine & SPP & 0.1235 & 0.0002 & 0.0665 & 0.1804 & Yes \\

\midrule
\midrule

\multirowcell{8}[-0.0ex][c]{\rotatebox[origin=c]{90}{\textbf{Engagingness}}}
& \multirowcell{1}[-0.0ex][l]{\textbf{No Refine vs MARA (Ours)}} 
& No Refine & MARA (Ours) & -0.6771 & 0.000 & -0.7304 & -0.6238 & Yes \\

& \multirowcell{1}[-0.0ex][l]{\textbf{No Refine vs MADR}} 
& No Refine & MADR & 0.6168 & 0.000 & 0.5635 & 0.6701 & Yes \\

& \multirowcell{1}[-0.0ex][l]{\textbf{No Refine vs Self-Refine}} 
& No Refine & Self-Refine & -0.1030 & 0.0006 & -0.1563 & -0.0497 & Yes \\

& \multirowcell{1}[-0.0ex][l]{\textbf{No Refine vs SPP}} 
& No Refine & SPP & 0.3065 & 0.000 & 0.2532 & 0.3597 & Yes \\

& \multirowcell{1}[-0.0ex][l]{\textbf{MARA (Ours) vs MADR}} 
& MARA (Ours) & MADR & 1.2939 & 0.000 & 1.2406 & 1.3472 & Yes \\

& \multirowcell{1}[-0.0ex][l]{\textbf{MARA (Ours) vs Self-Refine}} 
& MARA (Ours) & Self-Refine & 0.7801 & 0.000 & 0.7268 & 0.8334 & Yes \\

& \multirowcell{1}[-0.0ex][l]{\textbf{MARA (Ours) vs SPP}} 
& MARA (Ours) & SPP & -0.3706 & 0.000 & -0.4239 & -0.3173 & Yes \\

& \multirowcell{1}[-0.0ex][l]{\textbf{Self-Refine vs SPP}} 
& Self-Refine & SPP & 0.4095 & 0.000 & 0.3562 & 0.4628 & Yes \\

\bottomrule
\end{tabular}
}

\end{table*}

\begin{table*}[t!]
\caption{\small Significance Testing for Coherence, Groundedness, Naturalness, and Engagingness (PersonaChat Dataset)}
\vspace{-0.1in}
\label{tab:significance_personachat}
\small
\centering
\resizebox{1.0\textwidth}{!}{
\renewcommand{\arraystretch}{0.975}

\begin{tabular}{lcccccccc}
\toprule
& \textbf{Comparison} & \textbf{Model 1} & \textbf{Model 2} & \textbf{Mean Diff.} & \textbf{P-adj} & \textbf{Lower} & \textbf{Upper} & \textbf{Significant?} \\
\midrule
\midrule
\multirowcell{8}[-0.0ex][c]{\rotatebox[origin=c]{90}{\textbf{Coherence}}} 
& \multirowcell{1}[-0.0ex][l]{\textbf{No Refine vs MARA (Ours)}} & No Refine & MARA (Ours) & -0.4572 & 0.000 & -0.5092 & -0.4053 & Yes \\

& \multirowcell{1}[-0.0ex][l]{\textbf{No Refine vs MADR}} 
& No Refine & MADR & -0.2786 & 0.000 & -0.3264 & -0.2309 & Yes \\

& \multirowcell{1}[-0.0ex][l]{\textbf{No Refine vs Self-Refine}} 
& No Refine & Self-Refine & 0.3962 & 0.000 & 0.3442 & 0.4481 & Yes \\

& \multirowcell{1}[-0.0ex][l]{\textbf{No Refine vs SPP}} 
& No Refine & SPP & 0.1141 & 0.0002 & 0.0621 & 0.1660 & Yes \\

& \multirowcell{1}[-0.0ex][l]{\textbf{MARA (Ours) vs MADR}} 
& MARA (Ours) & MADR & 0.7359 & 0.000 & 0.6839 & 0.7878 & Yes \\

& \multirowcell{1}[-0.0ex][l]{\textbf{MARA (Ours) vs Self-Refine}} 
& MARA (Ours) & Self-Refine & -0.0611 & 0.0204 & -0.1130 & -0.0091 & Yes \\

& \multirowcell{1}[-0.0ex][l]{\textbf{MARA (Ours) vs SPP}} 
& MARA (Ours) & SPP & -0.3432 & 0.000 & -0.3951 & -0.2912 & Yes \\

& \multirowcell{1}[-0.0ex][l]{\textbf{Self-Refine vs SPP}} 
& Self-Refine & SPP & -0.2821 & 0.000 & -0.3340 & -0.2301 & Yes \\

\midrule
\midrule

\multirowcell{8}[-0.0ex][c]{\rotatebox[origin=c]{90}{\textbf{Groundedness}}}
& \multirowcell{1}[-0.0ex][l]{\textbf{No Refine vs MARA (Ours)}} 
& No Refine & MARA (Ours) & -0.1406 & 0.000 & -0.1627 & -0.1185 & Yes \\

& \multirowcell{1}[-0.0ex][l]{\textbf{No Refine vs MADR}} 
& No Refine & MADR & 0.1325 & 0.000 & 0.1104 & 0.1546 & Yes \\

& \multirowcell{1}[-0.0ex][l]{\textbf{No Refine vs Self-Refine}} 
& No Refine & Self-Refine & 0.1037 & 0.000 & 0.0816 & 0.1258 & Yes \\

& \multirowcell{1}[-0.0ex][l]{\textbf{No Refine vs SPP}} 
& No Refine & SPP & 0.0215 & 0.0574 & -0.0006 & 0.0436 & No \\

& \multirowcell{1}[-0.0ex][l]{\textbf{MARA (Ours) vs MADR}} 
& MARA (Ours) & MADR & 0.2731 & 0.000 & 0.2510 & 0.2952 & Yes \\

& \multirowcell{1}[-0.0ex][l]{\textbf{MARA (Ours) vs Self-Refine}} 
& MARA (Ours) & Self-Refine & -0.0369 & 0.0019 & -0.0590 & -0.0148 & Yes \\

& \multirowcell{1}[-0.0ex][l]{\textbf{MARA (Ours) vs SPP}} 
& MARA (Ours) & SPP & -0.1191 & 0.000 & -0.1412 & -0.0970 & Yes \\

& \multirowcell{1}[-0.0ex][l]{\textbf{Self-Refine vs SPP}} 
& Self-Refine & SPP & -0.0822 & 0.000 & -0.1043 & -0.0601 & Yes \\

\midrule
\midrule

\multirowcell{8}[-0.0ex][c]{\rotatebox[origin=c]{90}{\textbf{Naturalness}}}
& \multirowcell{1}[-0.0ex][l]{\textbf{No Refine vs MARA (Ours)}} 
& No Refine & MARA (Ours) & -0.4273 & 0.000 & -0.4802 & -0.3744 & Yes \\

& \multirowcell{1}[-0.0ex][l]{\textbf{No Refine vs MADR}} 
& No Refine & MADR & 0.1728 & 0.000 & 0.1199 & 0.2257 & Yes \\

& \multirowcell{1}[-0.0ex][l]{\textbf{No Refine vs Self-Refine}} 
& No Refine & Self-Refine & 0.5169 & 0.000 & 0.4640 & 0.5698 & Yes \\

& \multirowcell{1}[-0.0ex][l]{\textbf{No Refine vs SPP}} 
& No Refine & SPP & 0.1673 & 0.000 & 0.1144 & 0.2202 & Yes \\

& \multirowcell{1}[-0.0ex][l]{\textbf{MARA (Ours) vs MADR}} 
& MARA (Ours) & MADR & 0.6000 & 0.000 & 0.5471 & 0.6529 & Yes \\

& \multirowcell{1}[-0.0ex][l]{\textbf{MARA (Ours) vs Self-Refine}} 
& MARA (Ours) & Self-Refine & 0.0896 & 0.0017 & 0.0367 & 0.1425 & Yes \\

& \multirowcell{1}[-0.0ex][l]{\textbf{MARA (Ours) vs SPP}} 
& MARA (Ours) & SPP & -0.2599 & 0.000 & -0.3128 & -0.2070 & Yes \\

& \multirowcell{1}[-0.0ex][l]{\textbf{Self-Refine vs SPP}} 
& Self-Refine & SPP & -0.3496 & 0.000 & -0.4025 & -0.2967 & Yes \\

\midrule
\midrule

\multirowcell{8}[-0.0ex][c]{\rotatebox[origin=c]{90}{\textbf{Engagingness}}}
& \multirowcell{1}[-0.0ex][l]{\textbf{No Refine vs MARA (Ours)}} 
& No Refine & MARA (Ours) & -0.9953 & 0.000 & -1.0695 & -0.9211 & Yes \\

& \multirowcell{1}[-0.0ex][l]{\textbf{No Refine vs MADR}} 
& No Refine & MADR & 0.2260 & 0.000 & 0.1518 & 0.3002 & Yes \\

& \multirowcell{1}[-0.0ex][l]{\textbf{No Refine vs Self-Refine}} 
& No Refine & Self-Refine & 0.7535 & 0.000 & 0.6793 & 0.8277 & Yes \\

& \multirowcell{1}[-0.0ex][l]{\textbf{No Refine vs SPP}} 
& No Refine & SPP & 0.5208 & 0.000 & 0.4466 & 0.5950 & Yes \\

& \multirowcell{1}[-0.0ex][l]{\textbf{MARA (Ours) vs MADR}} 
& MARA (Ours) & MADR & 1.2213 & 0.000 & 1.1471 & 1.2955 & Yes \\

& \multirowcell{1}[-0.0ex][l]{\textbf{MARA (Ours) vs Self-Refine}} 
& MARA (Ours) & Self-Refine & 0.2418 & 0.000 & 0.1676 & 0.3160 & Yes \\

& \multirowcell{1}[-0.0ex][l]{\textbf{MARA (Ours) vs SPP}} 
& MARA (Ours) & SPP & -0.4745 & 0.000 & -0.5487 & -0.4003 & Yes \\

& \multirowcell{1}[-0.0ex][l]{\textbf{Self-Refine vs SPP}} 
& Self-Refine & SPP & -0.2327 & 0.000 & -0.3069 & -0.1585 & Yes \\

\bottomrule
\end{tabular}
}
\end{table*}

\begin{table*}[t!]
\caption{\small Significance Testing for Coherence, Groundedness, Naturalness, and Engagingness (INSCIT Dataset)}
\vspace{-0.1in}
\label{tab:significance_inscit}
\small
\centering
\resizebox{1.0\textwidth}{!}{
\renewcommand{\arraystretch}{0.975}
\begin{tabular}{lcccccccc}
\toprule
& \textbf{Comparison} & \textbf{Model 1} & \textbf{Model 2} & \textbf{Mean Diff.} & \textbf{P-adj} & \textbf{Lower} & \textbf{Upper} & \textbf{Significant?} \\
\midrule
\midrule
\multirowcell{8}[-0.0ex][c]{\rotatebox[origin=c]{90}{\textbf{Coherence}}} 
& \multirowcell{1}[-0.0ex][l]{\textbf{No Refine vs MARA (Ours)}} & No Refine & MARA (Ours) & 0.0844 & 0.0013 & 0.034 & 0.1348 & Yes \\

& \multirowcell{1}[-0.0ex][l]{\textbf{No Refine vs MADR}} 
& No Refine & MADR & -0.2420 & 0.000 & -0.2925 & -0.1916 & Yes \\

& \multirowcell{1}[-0.0ex][l]{\textbf{No Refine vs Self-Refine}} 
& No Refine & Self-Refine & -0.2250 & 0.000 & -0.2754 & -0.1746 & Yes \\

& \multirowcell{1}[-0.0ex][l]{\textbf{No Refine vs SPP}} 
& No Refine & SPP & -0.0102 & 0.9529 & -0.0582 & 0.0379 & No \\

& \multirowcell{1}[-0.0ex][l]{\textbf{MARA (Ours) vs MADR}} 
& MARA (Ours) & MADR & -0.3264 & 0.000 & -0.3769 & -0.2760 & Yes \\

& \multirowcell{1}[-0.0ex][l]{\textbf{MARA (Ours) vs Self-Refine}} 
& MARA (Ours) & Self-Refine & -0.3094 & 0.000 & -0.3598 & -0.2590 & Yes \\

& \multirowcell{1}[-0.0ex][l]{\textbf{MARA (Ours) vs SPP}} 
& MARA (Ours) & SPP & -0.0946 & 0.0005 & -0.1426 & -0.0465 & Yes \\

& \multirowcell{1}[-0.0ex][l]{\textbf{MADR vs Self-Refine}} 
& MADR & Self-Refine & 0.0171 & 0.7681 & -0.0310 & 0.0651 & No \\

& \multirowcell{1}[-0.0ex][l]{\textbf{MADR vs SPP}} 
& MADR & SPP & 0.2319 & 0.000 & 0.1839 & 0.2799 & Yes \\

& \multirowcell{1}[-0.0ex][l]{\textbf{Self-Refine vs SPP}} 
& Self-Refine & SPP & 0.2148 & 0.000 & 0.1668 & 0.2629 & Yes \\

\midrule
\midrule

\multirowcell{8}[-0.0ex][c]{\rotatebox[origin=c]{90}{\textbf{Groundedness}}}
& \multirowcell{1}[-0.0ex][l]{\textbf{No Refine vs MARA (Ours)}} 
& No Refine & MARA (Ours) & 0.0765 & 0.000 & 0.0526 & 0.1005 & Yes \\

& \multirowcell{1}[-0.0ex][l]{\textbf{No Refine vs MADR}} 
& No Refine & MADR & -0.0525 & 0.0003 & -0.0765 & -0.0285 & Yes \\

& \multirowcell{1}[-0.0ex][l]{\textbf{No Refine vs Self-Refine}} 
& No Refine & Self-Refine & -0.0721 & 0.0001 & -0.0961 & -0.0481 & Yes \\

& \multirowcell{1}[-0.0ex][l]{\textbf{No Refine vs SPP}} 
& No Refine & SPP & -0.0244 & 0.0518 & -0.0491 & 0.0002 & No \\

& \multirowcell{1}[-0.0ex][l]{\textbf{MARA (Ours) vs MADR}} 
& MARA (Ours) & MADR & -0.1290 & 0.000 & -0.1530 & -0.1051 & Yes \\

& \multirowcell{1}[-0.0ex][l]{\textbf{MARA (Ours) vs Self-Refine}} 
& MARA (Ours) & Self-Refine & -0.1486 & 0.000 & -0.1726 & -0.1247 & Yes \\

& \multirowcell{1}[-0.0ex][l]{\textbf{MARA (Ours) vs SPP}} 
& MARA (Ours) & SPP & -0.1010 & 0.000 & -0.1256 & -0.0764 & Yes \\

& \multirowcell{1}[-0.0ex][l]{\textbf{MADR vs Self-Refine}} 
& MADR & Self-Refine & -0.0196 & 0.1391 & -0.0442 & 0.0050 & No \\

& \multirowcell{1}[-0.0ex][l]{\textbf{MADR vs SPP}} 
& MADR & SPP & 0.0281 & 0.0245 & 0.0034 & 0.0527 & Yes \\

& \multirowcell{1}[-0.0ex][l]{\textbf{Self-Refine vs SPP}} 
& Self-Refine & SPP & 0.0477 & 0.0006 & 0.0231 & 0.0723 & Yes \\

\midrule
\midrule

\multirowcell{8}[-0.0ex][c]{\rotatebox[origin=c]{90}{\textbf{Naturalness}}}
& \multirowcell{1}[-0.0ex][l]{\textbf{No Refine vs MARA (Ours)}} 
& No Refine & MARA (Ours) & 0.1052 & 0.0036 & 0.0365 & 0.1739 & Yes \\

& \multirowcell{1}[-0.0ex][l]{\textbf{No Refine vs MADR}} 
& No Refine & MADR & -0.2434 & 0.000 & -0.3160 & -0.1708 & Yes \\

& \multirowcell{1}[-0.0ex][l]{\textbf{No Refine vs Self-Refine}} 
& No Refine & Self-Refine & -0.1727 & 0.0001 & -0.2414 & -0.1039 & Yes \\

& \multirowcell{1}[-0.0ex][l]{\textbf{No Refine vs SPP}} 
& No Refine & SPP & -0.0752 & 0.0309 & -0.1439 & -0.0065 & Yes \\

& \multirowcell{1}[-0.0ex][l]{\textbf{MARA (Ours) vs MADR}} 
& MARA (Ours) & MADR & -0.1382 & 0.0004 & -0.2108 & -0.0656 & Yes \\

& \multirowcell{1}[-0.0ex][l]{\textbf{MARA (Ours) vs Self-Refine}} 
& MARA (Ours) & Self-Refine & -0.0675 & 0.0549 & -0.1362 & 0.0013 & No \\

& \multirowcell{1}[-0.0ex][l]{\textbf{MARA (Ours) vs SPP}} 
& MARA (Ours) & SPP & 0.0300 & 0.6209 & -0.0387 & 0.0987 & No \\

& \multirowcell{1}[-0.0ex][l]{\textbf{MADR vs Self-Refine}} 
& MADR & Self-Refine & 0.0707 & 0.0430 & -0.0018 & 0.1433 & No \\

& \multirowcell{1}[-0.0ex][l]{\textbf{MADR vs SPP}} 
& MADR & SPP & 0.1682 & 0.0001 & 0.0995 & 0.2369 & Yes \\

& \multirowcell{1}[-0.0ex][l]{\textbf{Self-Refine vs SPP}} 
& Self-Refine & SPP & 0.0975 & 0.0062 & 0.0287 & 0.1662 & Yes \\

\midrule
\midrule

\multirowcell{8}[-0.0ex][c]{\rotatebox[origin=c]{90}{\textbf{Engagingness}}}
& \multirowcell{1}[-0.0ex][l]{\textbf{No Refine vs MARA (Ours)}} 
& No Refine & MARA (Ours) & 0.4544 & 0.000 & 0.3586 & 0.5502 & Yes \\

& \multirowcell{1}[-0.0ex][l]{\textbf{No Refine vs MADR}} 
& No Refine & MADR & -0.3843 & 0.000 & -0.4802 & -0.2885 & Yes \\

& \multirowcell{1}[-0.0ex][l]{\textbf{No Refine vs Self-Refine}} 
& No Refine & Self-Refine & -0.0373 & 0.6615 & -0.1331 & 0.0586 & No \\

& \multirowcell{1}[-0.0ex][l]{\textbf{No Refine vs SPP}} 
& No Refine & SPP & 0.1880 & 0.0003 & 0.0981 & 0.2779 & Yes \\

& \multirowcell{1}[-0.0ex][l]{\textbf{MARA (Ours) vs MADR}} 
& MARA (Ours) & MADR & -0.8387 & 0.000 & -0.9346 & -0.7429 & Yes \\

& \multirowcell{1}[-0.0ex][l]{\textbf{MARA (Ours) vs Self-Refine}} 
& MARA (Ours) & Self-Refine & -0.4917 & 0.000 & -0.5875 & -0.3958 & Yes \\

& \multirowcell{1}[-0.0ex][l]{\textbf{MARA (Ours) vs SPP}} 
& MARA (Ours) & SPP & -0.2664 & 0.000 & -0.3563 & -0.1765 & Yes \\

& \multirowcell{1}[-0.0ex][l]{\textbf{MADR vs Self-Refine}} 
& MADR & Self-Refine & 0.3471 & 0.000 & 0.2512 & 0.4429 & Yes \\

& \multirowcell{1}[-0.0ex][l]{\textbf{MADR vs SPP}} 
& MADR & SPP & 0.5723 & 0.000 & 0.4824 & 0.6623 & Yes \\

& \multirowcell{1}[-0.0ex][l]{\textbf{Self-Refine vs SPP}} 
& Self-Refine & SPP & 0.2253 & 0.0001 & 0.1354 & 0.3152 & Yes \\

\bottomrule
\end{tabular}
}

\end{table*}

\newpage

\begin{table*}
    \caption{The prompt used in the full instantiation of MARA for responding agent.}
    \label{tab:prompt:responding_agent}
    \vspace{-0.1in}
    \resizebox{0.95\textwidth}{!}{
    \renewcommand{\arraystretch}{1.1}
    \renewcommand{\tabcolsep}{2.5mm}
        \begin{tabular}{ll}
        \toprule
        \multicolumn{1}{p{.18\textwidth}}{\textbf{Types}} & \makecell{\multicolumn{1}{p{.92\textwidth}}{\textbf{Texts}}} \\
        \midrule
        \multicolumn{1}{p{.18\textwidth}}{\textbf{System Message}} & 

                \makecell{
            \multicolumn{1}{p{.92\textwidth}}{As a \texttt{<role>Responding Agent</role>}, your task is to answer the user's question, within the \texttt{<question\_text>} tags.} \\
            \multicolumn{1}{p{.92\textwidth}}{- Consider the Keyword: \texttt{<keyword>\{keyword\}</keyword>}, if available.} \\
            \multicolumn{1}{p{.92\textwidth}}{\texttt{<instructions>}} \\
            \multicolumn{1}{p{.92\textwidth}}{- Place your final response within \texttt{<response></response>} tags.} \\
            \multicolumn{1}{p{.92\textwidth}}{- Make your response concise.} \\
            \multicolumn{1}{p{.92\textwidth}}{\texttt{</instructions>}} \\
        } \\

        \midrule
        \multicolumn{1}{p{.18\textwidth}}{\textbf{User Message}} & 
        \makecell{
            \multicolumn{1}{p{.92\textwidth}}{\texttt{<question\_text>\{user\_query\}</question\_text>}} 
        } \\
        \bottomrule
        \end{tabular}
    }
\end{table*}

\begin{table*}
    \caption{The prompt used in the full instantiation of MARA for the planner agent.}
    \label{tab:prompt:planner_agent}
    \vspace{-0.1in}
    \resizebox{0.95\textwidth}{!}{
    \renewcommand{\arraystretch}{1.1}
    \renewcommand{\tabcolsep}{2.5mm}
        \begin{tabular}{ll}
        \toprule
        \multicolumn{1}{p{.18\textwidth}}{\textbf{Types}} & \makecell{\multicolumn{1}{p{.92\textwidth}}{\textbf{Texts}}} \\
        \midrule
        \multicolumn{1}{p{.18\textwidth}}{\textbf{System Message}} & 
        \makecell{
                \multicolumn{1}{p{.92\textwidth}}{Role: \texttt{<role>Planner Agent</role>}. Your task is to select the appropriate Agent(s) to refine the response to the user's query, following a step-by-step reasoning process.} \\
                \multicolumn{1}{p{.92\textwidth}}{\texttt{<considerations>}} \\
                \multicolumn{1}{p{.92\textwidth}}{- \texttt{<userProfile>\{persona\}</userProfile>}: Understand the user's profile to tailor your response.} \\
                \multicolumn{1}{p{.92\textwidth}}{- \texttt{<keywords>\{keyword\}</keywords>}: Identify the key topics and context of the conversation.} \\
                \multicolumn{1}{p{.92\textwidth}}{\texttt{</considerations>}} \\
                \multicolumn{1}{p{.92\textwidth}}{\texttt{<instructions>}} \\
                \multicolumn{1}{p{.92\textwidth}}{- Begin by examining the user's interests and the conversation's context as detailed in the \texttt{<userProfile>} and \texttt{<keywords>} tags. Also review the user's query in the \texttt{<question\_text>} tags and the initial response in the \texttt{<initialResponse>} tags.} \\
                \multicolumn{1}{p{.92\textwidth}}{- Next, read the roles of the three Refining Agents available, each tasked with specific aspects of response refinement: } \\
                \multicolumn{1}{p{.92\textwidth}}{\texttt{<Persona Refining Agent>} Verifies and refines the alignment with the user's profile and interests. \texttt{</Persona Refining Agent>}} \\
                \multicolumn{1}{p{.92\textwidth}}{\texttt{<Coherence Refining Agent>} Verifies and refines the coherence of responses. \texttt{</Coherence Refining Agent>}} \\
                \multicolumn{1}{p{.92\textwidth}}{\texttt{<Fact Refining Agent>} Verifies and refines the factual accuracy of responses. \texttt{</Fact Refining Agent>}} \\
                \multicolumn{1}{p{.92\textwidth}}{- Ensure that the response to the user's query in the \texttt{<question\_text>} tags is factually accurate, aligns with the user's profile, and is coherent.} \\
                \multicolumn{1}{p{.92\textwidth}}{- Most importantly, ensure the response fully addresses the user's question inside the \texttt{<question\_text>} tags.} \\
                \multicolumn{1}{p{.92\textwidth}}{- Based on your analysis of each Agent's contributions, decide which Agent or combination of Agents is best suited to refine the response in the \texttt{<agents\_set>} tag.} \\
                \multicolumn{1}{p{.92\textwidth}}{- Determine the optimal sequence of Agent involvement based on the necessary refinements. Write the order in \texttt{<agents\_set>}, separating each agent with a comma character (', '). Articulate your reasoning in the \texttt{<agents\_set\_justification>} tags.} \\
                \multicolumn{1}{p{.92\textwidth}}{- Also note that the sequence should reflect the priorities in refining the response to make it as relevant and accurate as possible. Also include the justification of the order in \texttt{<agents\_set\_order\_justification>}.} \\
                \multicolumn{1}{p{.92\textwidth}}{- If the initial response is sufficient and no further refinement is needed, write 'None' in the \texttt{<agents\_set>} tag.} \\
                \multicolumn{1}{p{.92\textwidth}}{- Place \texttt{<agents\_set>}, \texttt{<agents\_set\_order\_justification>}, and \texttt{<agents\_set\_justification>} as your final response in \texttt{<agent\_planning>} tags.} \\
                \multicolumn{1}{p{.92\textwidth}}{\texttt{</instructions>}} \\
            } \\ \\

        \midrule
        \multicolumn{1}{p{.18\textwidth}}{\textbf{User Message}} & 
        \makecell{
            \multicolumn{1}{p{.92\textwidth}}{{\texttt{<question\_text>\{user\_query\}</question\_text>} }} \\
            \multicolumn{1}{p{.92\textwidth}}{{\texttt{<initialResponse>\{initial\_response\}</initialResponse>}}} \\
        } \\
        \bottomrule
        \end{tabular}
    }
\end{table*}

\begin{table*}
    \caption{The prompt used in the full instantiation of MARA for fact-refining agent.}
    \label{tab:prompt:fact_refining_agent}
    \vspace{-0.1in}
    \resizebox{0.95\textwidth}{!}{
    \renewcommand{\arraystretch}{1.1}
    \renewcommand{\tabcolsep}{2.5mm}
        \begin{tabular}{ll}
        \toprule
        \multicolumn{1}{p{.18\textwidth}}{\textbf{Types}} & \makecell{\multicolumn{1}{p{.92\textwidth}}{\textbf{Texts}}} \\
        \midrule
        \multicolumn{1}{p{.18\textwidth}}{\textbf{System Message}} & 
    \makecell{
        \multicolumn{1}{p{.92\textwidth}}{
            Role: \texttt{<role>Fact Refining Agent</role>}. Your task is to refine the previous responses to ensure they are accurate within the context of the conversation, following a step-by-step reasoning process.
        } \\
        \multicolumn{1}{p{.92\textwidth}}{
            Consider the given topics (keywords) for the conversation, if available, marked by \texttt{<keywords>} tags: \texttt{<keywords>\{keyword\}</keywords>}.
        } \\
        \multicolumn{1}{p{.92\textwidth}}{
            \texttt{<instructions>}
        } \\
        \multicolumn{1}{p{.92\textwidth}}{
            - Begin by reviewing the keywords in the \texttt{<keywords>} tags to understand the context of the conversation. Note that we assume that the user's opening question specifically focuses on the keywords listed within the \texttt{<keywords>} tags. Therefore, do not request additional detail or clarification.
        } \\
        \multicolumn{1}{p{.92\textwidth}}{
            - First, examine the factual accuracy of the previous response provided in the \texttt{<factChecking>} tags.
        } \\
        \multicolumn{1}{p{.92\textwidth}}{
            - Document your verification outcome: If the response is factually accurate, place `Fact is verified.' inside the \texttt{<verification>} tags. If not, place `Fact is not verified.' in the \texttt{<verification>} tags.
        } \\
        \multicolumn{1}{p{.92\textwidth}}{
            - Specify the reasons for the verification in the \texttt{<verification\_justification>} tags.
        } \\
        \multicolumn{1}{p{.92\textwidth}}{
            - Next, refine the previous response to enhance factual correctness in the \texttt{<refined\_response>} tags. Describe each change you make and justify it based on factual accuracy.
        } \\
        \multicolumn{1}{p{.92\textwidth}}{
            - Place the refined response in the \texttt{<refined\_response>} tags and detail your reasoning for each refinement step in \texttt{<refinement\_justification>} tags.
        } \\
        \multicolumn{1}{p{.92\textwidth}}{
            - Finally, compile the \texttt{<verification>}, \texttt{<verification\_justification>}, \texttt{<refined\_response>} and \texttt{<refinement\_justification>} into the \texttt{<response>} tags, ensuring a clear and logical flow of thought.
        } \\
        \multicolumn{1}{p{.92\textwidth}}{
            - Maintain professionalism and avoid including apologies or acknowledgements in your response.
        } \\
        \multicolumn{1}{p{.92\textwidth}}{
            - Since this refined response is displayed directly to the user, address them as if you are the Responding Agent, not a Refining Agent. Avoid any reference to refining roles, the refinement process, or acknowledgment of a previous Refining Agent.
        } \\
        \multicolumn{1}{p{.92\textwidth}}{
            - Ensure your response shows you have considered the user's input inside the \texttt{<question\_text>} tags and their current state of mind. Keep your \texttt{<refined\_response>} concise, clear, and similar to human-written text.
        } \\
        \multicolumn{1}{p{.92\textwidth}}{
            \texttt{</instructions>}
        } \\
    } \\
        \midrule
        \multicolumn{1}{p{.18\textwidth}}{\textbf{User Message}} & 
        \makecell{
            \multicolumn{1}{p{.92\textwidth}}{
                This is the sequence of multiple agents involved in refining the initial response: \texttt{\{planned\_agent\_order\}}.
            } \\
            \multicolumn{1}{p{.92\textwidth}}{
                This is the justification for requiring multiple agents: \texttt{\{planned\_agents\_set\_justification\}}.
            } \\
            \multicolumn{1}{p{.92\textwidth}}{
                This is the justification for this sequence: \texttt{\{planned\_agent\_order\_justification\}}.
            } \\
            \multicolumn{1}{p{.92\textwidth}}{
                Note that you are a Fact Refining Agent.
            } \\
            \multicolumn{1}{p{.92\textwidth}}{
                This is the user's question, which should be fully addressed: \texttt{<question\_text>\{user\_query\}</question\_text>}
            } \\
            \multicolumn{1}{p{.92\textwidth}}{
                This is the initial response generated by the Responding Agent: \texttt{<initialResponse>\{initial\_response\}</initialResponse>}.
            } \\
            \multicolumn{1}{p{.92\textwidth}}{
                This is the refined response generated by the previous refining agent (\texttt{\{previous\_agent\_name\}}): \texttt{\{generated\_response\}}.
                } \\
            } \\
        \bottomrule
        \end{tabular}
    }
\end{table*}

\begin{table*}
    \caption{The prompt used in the full instantiation of MARA for persona-refining agent.}
    \label{tab:prompt:persona_refining_agent}
    \vspace{-0.1in}
    \resizebox{0.95\textwidth}{!}{
    \renewcommand{\arraystretch}{1.1}
    \renewcommand{\tabcolsep}{2.5mm}
        \begin{tabular}{ll}
        \toprule
        \multicolumn{1}{p{.18\textwidth}}{\textbf{Types}} & \makecell{\multicolumn{1}{p{.92\textwidth}}{\textbf{Texts}}} \\
        \midrule
        \multicolumn{1}{p{.18\textwidth}}{\textbf{System Message}} & 
    \makecell{
        \multicolumn{1}{p{.92\textwidth}}{
            Role: \texttt{<role>Persona Refining Agent</role>}. Your task involves verifying and refining responses based on the user's profile and interests. You will first verify the alignment and then refine the response if necessary, explaining each step in your reasoning.
        } \\
        \multicolumn{1}{p{.92\textwidth}}{
            Consider the User Profile: \texttt{<userProfile>\{persona\}</userProfile>}
        } \\
        \multicolumn{1}{p{.92\textwidth}}{
            \texttt{<instructions>}
        } \\
        \multicolumn{1}{p{.92\textwidth}}{
            - Begin by understanding the user's interests from the \texttt{<userProfile>} tags. 
        } \\
        \multicolumn{1}{p{.92\textwidth}}{
            - First, verify the alignment of the previous response inside the \texttt{<persona>} tags aligns with the user's preferences detailed in the \texttt{<userProfile>} tags. Explain your reasoning: If it aligns well, explain why you believe this alignment exists; if it does not, explain what aspects are misaligned or contradictory.
        } \\
        \multicolumn{1}{p{.92\textwidth}}{
            - Document your verification outcome: If the previous response is well-aligned, place `Persona is verified.' inside the \texttt{<verification>} tags. If not, place `Persona is not verified.' inside the \texttt{<verification>} tags.
        } \\
        \multicolumn{1}{p{.92\textwidth}}{
            - Specify the reasons for the verification in the \texttt{<verification\_justification>} tags.
        } \\
        \multicolumn{1}{p{.92\textwidth}}{
            - Next, refine the previous response to better align with the user's preferences detailed in the \texttt{<userProfile>} tags.
        } \\
        \multicolumn{1}{p{.92\textwidth}}{
            - Place the refined response in the \texttt{<refined\_response>} tags and detail your reasoning for each refinement step in the \texttt{<refinement\_justification>} tags.
        } \\
        \multicolumn{1}{p{.92\textwidth}}{
            - Finally, compile the \texttt{<verification>}, \texttt{<verification\_justification>}, \texttt{<refined\_response>} and \texttt{<refinement\_justification>} into the \texttt{<response>} tags, ensuring a clear and logical flow of thought.
        } \\
        \multicolumn{1}{p{.92\textwidth}}{
            - Maintain professionalism and avoid including apologies or acknowledgements in your response.
        } \\
        \multicolumn{1}{p{.92\textwidth}}{
            - Since this refined response is displayed directly to the user, address them as if you are the Responding Agent, not a Refining Agent. Avoid any reference to refining roles, the refinement process, or acknowledgment of a previous Refining Agent.
        } \\
        \multicolumn{1}{p{.92\textwidth}}{
            - Ensure your response shows you have considered the user's input inside the \texttt{<question\_text>} tags and their current state of mind. Keep your \texttt{<refined\_response>} concise, clear, and similar to human-written text.
        } \\
        \multicolumn{1}{p{.92\textwidth}}{
            \texttt{</instructions>}
        } \\
    } \\
        \midrule
        \multicolumn{1}{p{.18\textwidth}}{\textbf{User Message}} & 
        \makecell{
            \multicolumn{1}{p{.92\textwidth}}{
                This is the sequence of multiple agents involved in refining the initial response: \texttt{\{planned\_agent\_order\}}.
            } \\
            \multicolumn{1}{p{.92\textwidth}}{
                This is the justification for requiring multiple agents: \texttt{\{planned\_agents\_set\_justification\}}.
            } \\
            \multicolumn{1}{p{.92\textwidth}}{
                This is the justification for this sequence: \texttt{\{planned\_agent\_order\_justification\}}.
            } \\
            \multicolumn{1}{p{.92\textwidth}}{
                Note that you are a Persona Refining Agent.
            } \\
            \multicolumn{1}{p{.92\textwidth}}{
                This is the user's question, which should be fully addressed: \texttt{<question\_text>\{user\_query\}</question\_text>}
            } \\
            \multicolumn{1}{p{.92\textwidth}}{
                This is the initial response generated by the Responding Agent: \texttt{<initialResponse>\{initial\_response\}</initialResponse>}.
            } \\
            \multicolumn{1}{p{.92\textwidth}}{
                This is the refined response generated by the previous refining agent (\texttt{\{previous\_agent\_name\}}): \texttt{\{generated\_response\}}.
                } \\
            } \\
        \bottomrule
        \end{tabular}
    }
\end{table*}

\begin{table*}
    \caption{The prompt used in the full instantiation of MARA for coherence-refining agent.}
    \label{tab:prompt:coherence_refining_agent}
    \vspace{-0.1in}
    \resizebox{0.95\textwidth}{!}{
    \renewcommand{\arraystretch}{1.1}
    \renewcommand{\tabcolsep}{2.5mm}
        \begin{tabular}{ll}
        \toprule
        \multicolumn{1}{p{.18\textwidth}}{\textbf{Types}} & \makecell{\multicolumn{1}{p{.92\textwidth}}{\textbf{Texts}}} \\
        \midrule
        \multicolumn{1}{p{.18\textwidth}}{\textbf{System Message}} & 
        \makecell{
            \multicolumn{1}{p{.92\textwidth}}{
                Role: \texttt{<role>Coherence Refining Agent</role>}. Your task is to verify and refine the previous responses to ensure the coherence and logical flow of the initial responses within the context of the ongoing conversation and mimicking human-written text, following a step-by-step reasoning process.
            } \\
            \multicolumn{1}{p{.92\textwidth}}{
                Consider the given topics (keywords) for the conversation, if available, marked by \texttt{<keywords>} tags: \texttt{<keywords>\{keyword\}</keywords>}.
            } \\
            \multicolumn{1}{p{.92\textwidth}}{
                \texttt{<instructions>}
            } \\
            \multicolumn{1}{p{.92\textwidth}}{
                - Begin by reviewing the keywords in the \texttt{<keywords>} tags to understand the context of the conversation. Note that we assume that the user's opening question specifically focuses on the keywords listed within the \texttt{<keywords>} tags. Therefore, do not request additional detail or clarification.
            } \\
            \multicolumn{1}{p{.92\textwidth}}{
                - First, examine whether the previous response provided in the \texttt{<coherence>} tags maintains coherence with the conversation history, mimicking human-written text.
            } \\
            \multicolumn{1}{p{.92\textwidth}}{
                - Document your verification outcome: If the response is coherent, place `Coherence is verified.' inside the \texttt{<verification>} tags. If not, place `Coherence is not verified.' in the \texttt{<verification>} tags.
            } \\
            \multicolumn{1}{p{.92\textwidth}}{
                - Specify the reasons for the verification in the \texttt{<verification\_justification>} tags.
            } \\
            \multicolumn{1}{p{.92\textwidth}}{
                - Next, refine the previous response to improve the overall clarity and continuity of the conversation in the \texttt{<refined\_response>} tags. Describe each change you make and justify it based on coherence and completeness.
            } \\
            \multicolumn{1}{p{.92\textwidth}}{
                - Place the refined response in the \texttt{<refined\_response>} tags and detail your reasoning for each refinement step in \texttt{<refinement\_justification>} tags.
            } \\
            \multicolumn{1}{p{.92\textwidth}}{
                - Finally, compile the \texttt{<verification>}, \texttt{<verification\_justification>}, \texttt{<refined\_response>} and \texttt{<refinement\_justification>} into the \texttt{<response>} tags, ensuring a clear and logical flow of thought.
            } \\
            \multicolumn{1}{p{.92\textwidth}}{
                - Maintain professionalism and avoid including apologies or acknowledgements in your response.
            } \\
            \multicolumn{1}{p{.92\textwidth}}{
                - Since this refined response is displayed directly to the user, address them as if you are the Responding Agent, not a Refining Agent. Avoid any reference to refining roles, the refinement process, or acknowledgment of a previous Refining Agent.
            } \\
            \multicolumn{1}{p{.92\textwidth}}{
                - Ensure your response shows you have considered the user's input inside the \texttt{<question\_text>} tags and their current state of mind. Keep your \texttt{<refined\_response>} concise, clear, and similar to human-written text.
            } \\
            \multicolumn{1}{p{.92\textwidth}}{
                \texttt{</instructions>}
            } \\
        } \\
        \midrule
        \multicolumn{1}{p{.18\textwidth}}{\textbf{User Message}} & 
        \makecell{
            \multicolumn{1}{p{.92\textwidth}}{
                This is the sequence of multiple agents involved in refining the initial response: \texttt{\{planned\_agent\_order\}}.
            } \\
            \multicolumn{1}{p{.92\textwidth}}{
                This is the justification for requiring multiple agents: \texttt{\{planned\_agents\_set\_justification\}}.
            } \\
            \multicolumn{1}{p{.92\textwidth}}{
                This is the justification for this sequence: \texttt{\{planned\_agent\_order\_justification\}}.
            } \\
            \multicolumn{1}{p{.92\textwidth}}{
                Note that you are a Coherence Refining Agent.
            } \\
            \multicolumn{1}{p{.92\textwidth}}{
                This is the user's question, which should be fully addressed: \texttt{<question\_text>\{user\_query\}</question\_text>}
            } \\
            \multicolumn{1}{p{.92\textwidth}}{
                This is the initial response generated by the Responding Agent: \texttt{<initialResponse>\{initial\_response\}</initialResponse>}.
            } \\
            \multicolumn{1}{p{.92\textwidth}}{
                This is the refined response generated by the previous refining agent (\texttt{\{previous\_agent\_name\}}): \texttt{\{generated\_response\}}.
                } \\
            } \\
        \bottomrule
        \end{tabular}
    }
\end{table*}
\begin{table*}
    \caption{The prompt template used for G-Eval (Coherence).}
    \label{tab:prompt:geval_coherence}
    \vspace{-0.1in}
    \renewcommand{\arraystretch}{1.2}
    \begin{tabular}{p{0.97\textwidth}}
    \toprule
    \textbf{Instructions:} \\
    You will be given a conversation segment involving two participants: the user and the system. You will then be given one potential response for the next turn in the conversation. Your task is to rate the generated responses on one metric. Please make sure you read and understand these instructions carefully. Please keep this document open while reviewing, and refer to it as needed. \\
    \midrule
    \textbf{Evaluation Criteria:} \\
    \textbf{Coherence (1-3)} - Evaluate whether the conversation response logically follows the preceding context and maintains a clear, logical flow. \\
    \textbf{Score of 1:} Assign this score if the response does not logically follow the preceding context. It may introduce abrupt changes in topic or contain confusing statements, resulting in a disjointed conversation. \\
    \textbf{Score of 2:} Assign this score if the response somewhat follows the preceding context but includes minor logical inconsistencies or slight topic shifts, slightly disrupting the conversation's flow. \\
    \textbf{Score of 3:} Assign this score if the response logically follows the preceding context and maintains a clear, logical flow, providing a relevant and coherent continuation of the conversation without any abrupt changes or confusing elements. \\
    \midrule
    \textbf{Evaluation Steps:} \\
    1. Examine the conversational history, the provided fact (if given), and the user profile (if given) to fully understand the context and dynamics of the conversation. \\
    2. Compare the generated response with the gold standard response to evaluate how well it maintains continuity and logical flow. \\
    3. Assess how effectively the generated response connects with and continues the conversation, ensuring it aligns logically with the existing conversation. \\
    4. Based on your analysis, assign a coherence score from 1 to 3, reflecting the response’s logical integration into the ongoing conversation. \\
    \midrule
    \textbf{Example:} \\
    Conversation History: \{\{Document\}\} \\
    Corresponding Fact: \{\{Fact\}\} \\
    Corresponding User Profile: \{\{Persona\}\} \\
    Gold Standard Response: \{\{Gold\_Response\}\} \\
    Generated Response: \{\{Response\}\} \\
    \midrule
    \textbf{Evaluation Form (Scores ONLY without any additional text):} \\
    - Coherence: \\
    \bottomrule
    \end{tabular}
\end{table*}

\begin{table*}
    \caption{The prompt template used for G-Eval (Groundedness).}
    \label{tab:prompt:geval_groundedness}
    \vspace{-0.1in}
    \renewcommand{\arraystretch}{1.2}
    \begin{tabular}{p{0.97\textwidth}}
    \toprule
    \textbf{Instructions:} \\
    You will be given a conversation segment involving two participants: the user and the system. You will then be given one potential response for the next turn in the conversation. The response concerns an interesting fact, which will be provided as well. Your task is to rate the generated responses on one metric. Please make sure you read and understand these instructions carefully. Please keep this document open while reviewing, and refer to it as needed. \\
    \midrule
    \textbf{Evaluation Criteria:} \\
    \textbf{Groundedness (0-1)} - Evaluate whether the conversation response is based on and accurately incorporates the provided fact. \\
    \textbf{Score of 0:} Assign this score if the response does not correctly use the provided fact or misrepresents it. This includes instances where the response contains inaccuracies or fails to integrate the fact meaningfully into the conversation. \\
    \textbf{Score of 1:} Assign this score if the response uses the provided fact accurately and integrates it seamlessly into the conversation, thereby enhancing the dialogue's relevance and informativeness. \\
    \midrule
    \textbf{Evaluation Steps:} \\
    1. Examine the conversational history and the provided fact (if given) to understand the context fully. \\
    2. Evaluate how accurately and relevantly the generated response incorporates the provided fact, ensuring there are no inaccuracies or hallucinated details. \\
    3. Analyze how the generated response measures up against a gold standard response to understand the ideal integration of the fact. \\
    4. Based on your assessment, assign a Groundedness score ranging from 0 to 1, reflecting how effectively the fact is incorporated into the response. \\
    \midrule
    \textbf{Example:} \\
    Conversation History: \{\{Document\}\} \\
    Corresponding Fact: \{\{Fact\}\} \\
    Gold Standard Response: \{\{Gold\_Response\}\} \\
    Generated Response: \{\{Response\}\} \\
    \midrule
    \textbf{Evaluation Form (Scores ONLY without any additional text):} \\
    - Groundedness: \\
    \bottomrule
    \end{tabular}
\end{table*}

\begin{table*}
    \caption{The prompt template used for G-Eval (Naturalness).}
    \label{tab:prompt:geval_naturalness}
    \vspace{-0.1in}
    \renewcommand{\arraystretch}{1.2}
    \begin{tabular}{p{0.97\textwidth}} 
    \toprule
    \textbf{Instructions} \\
    You will be given a conversation segment involving two participants: the user and the system. You will then be given one potential response for the next turn in the conversation. Your task is to rate the generated responses on one metric. Please make sure you read and understand these instructions carefully. Please keep this document open while reviewing, and refer to it as needed. \\
    \midrule
    \textbf{Evaluation Criteria:} \\
    \textbf{Naturalness (1-3)} - Evaluate whether the dialogue response feels natural and conversational, as if it were part of a real, human conversation. \\
    \textbf{Score of 1:} Assign this score if the response does not sound natural. It may contain awkward phrasing, unnatural expressions, or robotic language, disrupting the flow of the conversation. \\
    \textbf{Score of 2:} Assign this score if the response somewhat sounds natural but may include minor awkwardness or slightly unnatural phrasing, affecting the overall conversational flow. \\
    \textbf{Score of 3:} Assign this score if the response sounds completely natural, flowing smoothly, using natural language, and integrating seamlessly into the conversation as if it were part of a real human interaction. \\
    \midrule
    \textbf{Evaluation Steps:} \\
    1. Examine the conversational history, the provided fact (if given), and the user profile (if given) to gauge the natural fit of the response within the conversation’s context. \\
    2. Evaluate the tone, formality, and conversational flow of the generated response to determine how naturally it fits into the dialogue. \\
    3. Compare the generated response to a gold standard response to gauge the ideal level of naturalness. \\
    4. Based on your assessment, assign a Naturalness score from 1 to 3, focusing on how naturally the response fits into the conversation. \\
    \midrule
    \textbf{Example:} \\
    Conversation History: \{\{Document\}\} \\
    Corresponding Fact: \{\{Fact\}\} \\
    Corresponding User Profile: \{\{Persona\}\} \\
    Gold Standard Response: \{\{Gold\_Response\}\} \\
    Generated Response: \{\{Response\}\} \\
    \midrule
    \textbf{Evaluation Form (Scores ONLY without any additional text):} \\
    - Naturalness: \\
    \bottomrule
    \end{tabular}
\end{table*}

\begin{table*}
    \caption{The prompt template used for G-Eval (Engagingness).}
    \label{tab:prompt:geval_engagingness}
    \vspace{-0.1in}
    \renewcommand{\arraystretch}{1.2}
    \begin{tabular}{p{0.97\textwidth}}
    \toprule
    \textbf{Instructions} \\
    You will be given a conversation segment involving two participants: the user and the system. You will then be given one potential response for the next turn in the conversation. Your task is to rate the generated responses on one metric. Please make sure you read and understand these instructions carefully. Please keep this document open while reviewing, and refer to it as needed. \\
    \midrule
    \textbf{Evaluation Criteria:} \\
    \textbf{Engagingness (1-3)} - Is the response dull or interesting? \\
    \textbf{Score of 1 (Dull):} Assign this score if the response is generic and unremarkable, failing to spark interest or engagement. \\
    \textbf{Score of 2 (Somewhat Interesting):} Assign this score if the response is moderately interesting and could engage participants in the conversation, such as by introducing an opinion or thought. \\
    \textbf{Score of 3 (Interesting):} Assign this score if the response is highly interesting or presents an intriguing fact, significantly enhancing the conversation's appeal. \\
    \midrule
    \textbf{Evaluation Steps:} \\
    1. Examine the conversational history, the provided fact (if given), and the user profile (if given) to gauge the potential interest or intrigue. \\
    2. Assess how the generated response contributes to the conversation’s value and captivates interest. \\
    3. Compare the generated response to a gold standard response. \\
    4. Based on your analysis, assign an Engagingness score from 1 to 3, reflecting the response’s ability to captivate and add value to the conversation. \\
    \midrule
    \textbf{Example:} \\
    Conversation History: \{\{Document\}\} \\
    Corresponding Fact: \{\{Fact\}\} \\
    Corresponding User Profile: \{\{Persona\}\} \\
    Gold Standard Response: \{\{Gold\_Response\}\} \\
    Generated Response: \{\{Response\}\} \\
    \midrule
    \textbf{Evaluation Form (Scores ONLY without any additional text):} \\
    - Engagingness: \\
    \bottomrule
    \end{tabular}
\end{table*}

\end{document}